\documentclass{article}

\usepackage{arxiv}
\usepackage{commath}
\usepackage{amssymb}

\usepackage{subcaption}
\usepackage{booktabs} 
\usepackage{multirow}
\usepackage{amsthm}
\usepackage{mathtools, nccmath}
\usepackage{url}
\usepackage[dvipsnames]{xcolor}
\usepackage{cool}
\usepackage{algorithm}
\usepackage[utf8]{inputenc} 
\usepackage[T1]{fontenc}    
\usepackage{hyperref}       
\usepackage{url}            
\usepackage{booktabs}       
\usepackage{amsfonts}       
\usepackage{nicefrac}       
\usepackage{microtype}      
\usepackage{lipsum}		
\usepackage{graphicx}
\usepackage{natbib}
\usepackage{doi}
\usepackage{lscape} 
\usepackage{array}

\newcolumntype{P}[1]{>{\centering\arraybackslash}p{#1}}

\title{data-driven theory-guided learning of partial differential equations using SimultaNeous basis function Approximation and Parameter Estimation (\textbf{SNAPE})}

\author{ \href{https://orcid.org/0000-0001-9350-4803}{\includegraphics[scale=0.06]{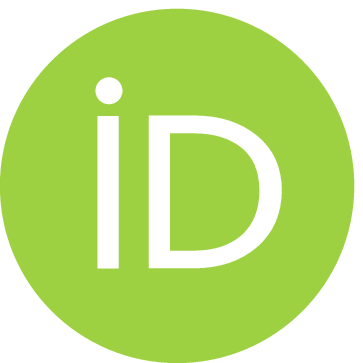}\hspace{1mm}Sutanu Bhowmick} \\
	Department of Civil and Environmental Engineering\\
	Rice University\\
	Houston, TX 77005 \\
	\texttt{sutanu.bhowmick@rice.edu} \\
	\And
	\href{https://orcid.org/0000-0003-0088-1656}{\includegraphics[scale=0.06]{orcid.eps}\hspace{1mm}Satish Nagarajaiah}\thanks{Corresponding author}\\
	Department of Civil and Environmental Engineering\\
	Department of Mechanical Engineering\\
	Rice University\\
	Houston, TX 77005 \\
	\texttt{satish.nagarajaiah@rice.edu} \\
}

\renewcommand{\headeright}{A Preprint}
\renewcommand{\undertitle}{A Preprint}
\renewcommand{\shorttitle}{\textbf{SNAPE}: Theory-guided learning of PDEs from data}

\hypersetup{
pdftitle={SNAPE-Theory-guided learning of PDEs from data},
pdfsubject={physics.class-ph, math.AP, math.DS, math.OC, cs.LG, nlin.CD},
pdfauthor={Sutanu Bhowmick, Satish Nagarajaiah},
pdfkeywords={Partial differential equations, Parameter estimation, Basis function approximation, Theory-guided learning, ADMM optimization},
}

\newcommand\blfootnote[1]{%
	\begingroup
	\renewcommand\thefootnote{}\footnote{#1}%
	\addtocounter{footnote}{-1}%
	\endgroup
}

\begin{document}
\maketitle

\begin{abstract}
	The measured spatiotemporal response of various physical processes is utilized to infer the governing partial differential equations (PDEs). We propose \textbf{S}imulta\textbf{N}eous Basis Function \textbf{A}pproximation and \textbf{P}arameter \textbf{E}stimation (\textbf{SNAPE}), a technique of parameter estimation of PDEs that is robust against high levels of noise nearly 100\%, by simultaneously fitting basis functions to the measured response and estimating the parameters of both ordinary and partial differential equations. The domain knowledge of the general multidimensional process is used as a constraint in the formulation of the optimization framework. SNAPE not only demonstrates its applicability on various complex dynamic systems that encompass wide scientific domains including Schrödinger equation, chaotic duffing oscillator, and Navier-Stokes equation but also estimates an analytical approximation to the process response. The method systematically combines the knowledge of well-established scientific theories and the concepts of data science to infer the properties of the process from the observed data.\blfootnote{\textit{Preprint submitted to International Journal}} 
\end{abstract}


\section{Introduction}
\label{sec:Intro}

Sensors measuring analog responses of a general multidimensional process at discrete spatial locations are becoming superior and more affordable \citet{tsang1985theory, zhu20203d, akyildiz2002wireless, badon2016smart, bhowmick2020measurement, bhowmick2022spatiotemporal, adrian1991particle, sun2015carbon, chu1985applications, yang2017full, yang2016dynamic}. Concurrently, the evolving big data storage facilities and computational capabilities can harness such high dimensional data  \citet{marx2013big, demchenko2013addressing, sun2020review} to inquire more about the underlying physical laws. Such physical laws have been extensively studied in the past to put forward scientific theories having mathematical formulations. Most often such well-studied scientific theories are represented in the form of ordinary or partial differential equations. In the last century, the research was directed towards forward-modeling which consists of obtaining analytical and numerical solutions of differential equations. With the advent of high-dimensional sensing systems and the acquired big data, recently the research is more focused on learning about the parameters of the continuous spatiotemporal process by addressing the inverse problem \citet{tarantola2006popper, tarantola2005inverse, lieberman2010parameter, nagarajaiah2017modeling}. As scientists and engineers, we are cognizant of the governing theory of the multidimensional analog processes that are measured digitally. The domain knowledge allows for the description of the physical process in the form of a mathematical model. But we need to estimate the unknown parameters that identify the final connection between the observations we are measuring and the inherent physical processes which characterize them. Several studies have been conducted previously to estimate the parameters of the ordinary differential equation (ODE) models from its observations \citep{ramsay2007parameter, peifer2007parameter, brunton2016discovering, lai2019sparse, lai2021structural}. The problem becomes harder in the case of models represented by partial differential equations (PDEs) compared to ODEs as the former includes differentials with respect to multiple variables depending on the dimensions of the model (e.g. spatiotemporal PDE models in fluid mechanics, wave optics, or geophysics).

One of the prevalent approaches of estimating PDE parameters involves optimizing the parameter space of the PDE by minimizing the difference between the numerically simulated response to the observed measurements \citep{muller2002fitting}. But the optimization problem suffers from the presence of local minima different from global minima (non-convex) \citep{muller2004parameter}. Also, the method requires knowledge of the boundary conditions and involves large computational cost. The other approach is based on regression analysis to estimate parameters of the temporal and spatial derivative terms in the PDE model \citep{bar1999fitting,voss1999amplitude,liang2008parameter}. The spatial and temporal derivatives are obtained from the measured process data by performing numerical differentiation. This two-stage approach of numerical differentiation and regression has been preferred over the first approach because of its computational simplicity. \citet{rudy2017data} and \citet{schaeffer2017learning} extend the two-stage method to discover the structure of the PDE model from an overcomplete dictionary of feasible mathematical terms by implementing sparse linear regression. \citet{xun2013parameter} extends the generalized smoothing approach of \citet{ramsay2007parameter} for ODE models to estimate the parameters of PDE models. In recent times, with the emergence of big data and high-performance computational frameworks, deep learning algorithms have been implemented to address inverse problems in diverse scientific fields such as biomedical imaging \citep{lucas2018using,ongie2020deep,jin2017deep}, geophysics \citep{seydoux2020clustering, zhang2019regularized}, cosmology \citep{ribli2019improved} to name a few. Similar attempts have been made to solve the inverse problem of PDE model identification by using deep neural networks \citep{raissi2019physics, long2018pde, long2019pde, both2021deepmod}. The general approach involves fitting the measured response variable using a deep regression neural network. A separate neural network enables the implementation of the PDE model using automatic/numerical differentiation of the fitted response model with respect to the independent variables.

The previously presented methods can be broadly categorized into two classes: regression-based and deep learning-based methods. Both classes of methods identify the latent PDE model from the measured full-field data devoid of the iterative numerical solution of the PDE model, thereby achieving higher computational efficiency. Nonetheless, both classes of methods suffer from significant drawbacks that (a) the regression-based method suffers from the inaccurate estimation of numerical derivatives in the presence of noise, especially the higher-order derivatives \citet{rudy2017data}, and (b) the deep learning methods lack any formal rule regarding the choice of network architecture, initialization, activation functions, or optimization schemes \citet{raissi2019physics}. The first limitation has been explicitly mentioned by the authors in \citep{rudy2017data} where they report a substantial error in the estimation of the parameter of a fourth-order Kuramoto-Sivashinsky PDE model in the presence of a small amount of noise. The second limitation is discussed in greater detail by \citep{raissi2019physics}. Not only the scientific interpretability of the deep learning models is absent \citet{gilpin2018explaining, ribeiro2016model}, but also there is growing skepticism over the stability of its solution to the inverse problems \citet{antun2020instabilities, gottschling2020troublesome}. The repeatability of its outcomes \citet{hutson2018artificial, vamathevan2019applications} on account of randomness in the data or initialization and its robustness against adversarial perturbations \citet{belthangady2019applications} have been increasingly questioned. Such concerns of repeatability can be found in the deep learning model of \citep{both2021deepmod} where the method identifies the PDE models only for some of the randomized trials.

This paper addresses the above-mentioned shortcomings by proposing the method of \textbf{SNAPE} (\textbf{S}imulta\textbf{N}eous Basis Function \textbf{A}pproximation and \textbf{P}arameter \textbf{E}stimation) which stands on the ideals of theory-guided learning \citep{karpatne2017theory, roscher2020explainable}, a progressive practice of data science in the scientific community. \textbf{SNAPE} infers the parameters of the linear and nonlinear differential equation (both ODEs and PDEs) models from the measured observations of the responses with the use of domain knowledge of the physical process or any general multidimensional processes. The proposed method in this paper incorporates the concept of a generalized smoothing approach by fitting basis functions to the measured response; unlike studies by \citep{ramsay2007parameter} and \citep{xun2013parameter} wherein discrete sampling of penalized splines is adopted. Such approximate numerical treatments are not amenable in noisy conditions and have to be replaced by exact differentiation. In this paper we propose the use of exact differentiation of spline basis functions. The coefficients of the basis functions are constrained to satisfy the differential equation for all the observed measurements of the multidimensional general process response. The parameters of the differential equations, as well as the coefficients of the basis functions, are simultaneously evaluated using the alternating direction method of multipliers (ADMM) optimization algorithm \citet{gabay1976dual, yang2011alternating, boyd2011distributed}. The proposed method does not require knowledge of the initial or boundary conditions of the model. \textbf{SNAPE} demonstrates its robustness by successfully estimating parameters of differential equation models from data perturbed with a large amount of noise (nearly 100\% Gaussian noise). The repeatability of the proposed method is guaranteed by inferring the model parameters from parametric bootstrap samples \citet{efron1994introduction}, thereby obtaining the mean and the confidence bounds of the estimates.

\section{Results}
\label{sec:Results}

\begin{figure}
	\centering
	\includegraphics[width=1.0\textwidth]{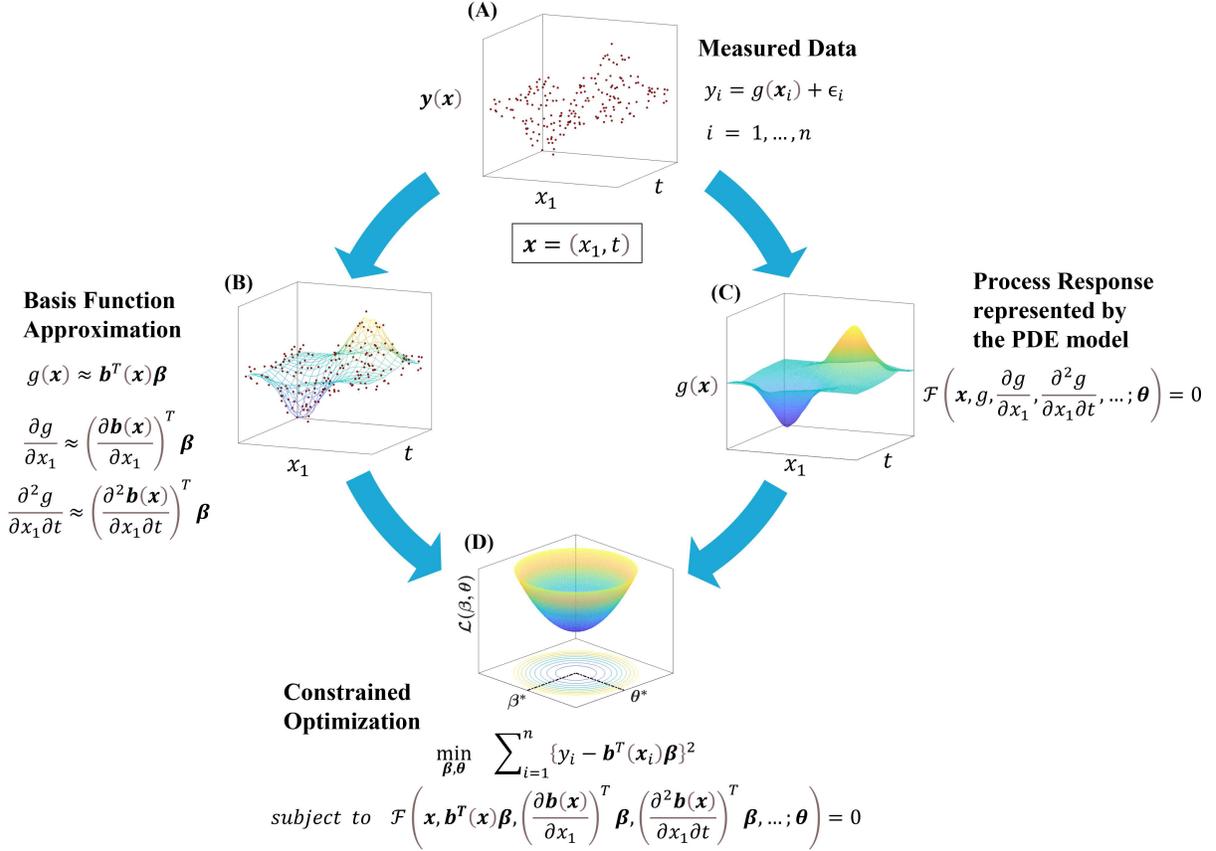}
	\caption{\textbf{SimultaNeous basis function Approximation and Parameter Estimation (SNAPE) of Partial Differential Equation (PDE) models from data.} \textbf{(A) }Measured data with noise $\mathrm{\epsilon}$ of a general two-dimensional dynamic process $g\left(\mathbf{x}\right)$ at $n$ discrete points \textbf{ (B) }basis function approximation with unknown coefficients $\boldsymbol{\beta }$ of both the process response and its partial derivatives with respect to the independent variables $\mathbf{x}$ \textbf{ (C) }the PDE model $\mathcal{F}\left(\right)=\mathbf{0}$ as a function of the independent variables, the response, and its partial derivatives with unknown parameters $\boldsymbol{\theta }$, which governs the process  \textbf{(D)} the response and the partial derivative terms of the PDE model are constrained to simultaneously obtain the optimum basis coefficients ${\boldsymbol{\beta }}^*$ that approximates the measured response and the optimum parameters ${\boldsymbol{\theta }}^*$ that satisfy the underlying PDE model by the measured data. Likewise, the application of \textbf{SNAPE} algorithm proposed herein can be adopted for PDE models by generalizing to any multidimensional processes (i.e., $\mathbf{x}\mathrm{=}\left(x_{\mathrm{1}},x_{\mathrm{2}}\mathrm{,\dots }t\right)\mathrm{\in }{\mathbb{R}}^p$).}\label{fig:1}
\end{figure}

A multidimensional dynamic process is represented by its response $g\left(\mathbf{x}\right)$, with $\mathbf{x}\mathrm{=}\left(x_{\mathrm{1}},x_{\mathrm{2}}\mathrm{,\dots }t\right)\mathrm{\in }{\mathbb{R}}^p$ being the multidimensional domain of the process. In the case of solid and fluid mechanics, the domain may consist of three spatial and one temporal coordinate. In the subsequent part, the application of the proposed method is described using PDEs that provide a more generalized form of a differential equation. Such initial-boundary value problems are represented by a PDE model which is satisfied within the domain $\mathbf{x}\mathrm{\in }\mathbf{\boldsymbol{\varOmega}}$  given by

\begin{equation} \label{EQ01} 
	\mathcal{F}\left(\mathbf{x},g,\dots ,g^q,\frac{\partial g}{\partial x_1},\dots ,\frac{\partial g}{\partial t},\frac{{\partial }^2g}{\partial x_1\partial x_1},\dots ,\frac{{\partial }^2g}{\partial x_1\partial t},g\frac{\partial g}{\partial x_1},\dots ,\frac{\partial g}{\partial x_1}\frac{\partial g}{\partial t},\dots ;\boldsymbol{\theta }\right)=0,\qquad \mathbf{x}\in \boldsymbol{\varOmega} 
\end{equation} 

where the parameter vector $\boldsymbol{\theta }=\left({\theta }_1,\dots ,{\theta }_m\right)$ are the coefficients of the PDE model having parametric form in $g\left(\mathbf{x}\right)$ and its partial derivatives. The uniqueness of the solution is established by defining the initial and boundary conditions of the aforementioned process which is satisfied at the boundary of the domain $\mathbf{x}\mathrm{\in }\boldsymbol{\varGamma}$ given by  

\begin{equation} \label{EQ02} 
	\mathcal{H}\left(g,\dots ,g^q,\frac{\mathrm{\partial }g}{\mathrm{\partial }x_1},\dots ,\frac{\mathrm{\partial }g}{\mathrm{\partial }t},\dots ,\frac{{\mathrm{\partial }}^qg}{\mathrm{\partial }x^q_1},\dots ,\frac{{\mathrm{\partial }}^qg}{\mathrm{\partial }t^q},\dots \right)=h\left(\mathbf{x}\right),\qquad \mathbf{x}\mathrm{\in }\boldsymbol{\varGamma} 
\end{equation} 

The initial or the boundary conditions are referred to as homogeneous if $h\left(\mathbf{x}\right)=0$. The PDE model in equations \ref{EQ01} and \ref{EQ02} represents the most general form of constant-coefficient nonlinear PDE model of arbitrary order. Even if the solution of the PDE model represents continuous multivariate function and its domain $\boldsymbol{\varOmega}$ and boundary $\boldsymbol{\varGamma}$ represents continuous functional space, in a practical scenario we acquire data in discrete points of the multidimensional domain which are contaminated with measurement noise. Assuming $g\left(\mathbf{x}\right)$ is measured as its surrogate $\mathbf{y}\left(\mathbf{x}\right)$ at discrete points within the multidimensional domain $\boldsymbol{\varOmega}$, $\mathbf{x}\mathrm{=}\left(x_{\mathrm{1}},x_{\mathrm{2}}\mathrm{,\dots }t\right)\ \mathrm{\in }{\mathbb{R}}^p$ having the measurements $\left(y_i,{\mathbf{x}}_i\right)$, where $i\ =\ 1,\dots ,n$ satisfying $y_i=g\left({\mathbf{x}}_i\right)+{\mathrm{\epsilon}}_i$. The independent and identically distributed homoscedastic measurement noise ${\mathrm{\epsilon}}_i$, $i\ =\ 1,\dots ,n$ are assumed to follow a Gaussian distribution with zero mean and ${\mathrm{\sigma}}^{\mathrm{2}}_{\mathrm{\epsilon}}$ variance.

The objective of the present study is to estimate the unknown $\boldsymbol{\theta }$ in the PDE model of equation \ref{EQ01} from the noisy measurement data. The proposed method of \textbf{SNAPE} takes into account the PDE model and the associated unknown parameter vector $\boldsymbol{\theta }=\left({\theta }_1,\dots ,{\theta }_m\right)$ by expressing the process response $g\left(\mathbf{x}\right)$ as an approximation to the linear combination of basis functions given by

\begin{equation} \label{EQ03} 
	g\left(\mathbf{x}\right)\approx \overline{g}\left(\mathbf{x}\right)=\sum^K_{k=1}{b_k\left(\mathbf{x}\right){\beta }_k}={\mathbf{b}}^T\left(\mathbf{x}\right)\boldsymbol{\beta } 
\end{equation}

where $\mathbf{b}\left(\mathbf{x}\right)=\{b_1\left(\mathbf{x}\right),\dots ,b_K\left(\mathbf{x}\right){\}}^T$ is the vector of basis functions and $\boldsymbol{\beta }={\left({\beta }_1,\dots ,{\beta }_K\right)}^T$ is the vector of basis coefficients. In this study, the B-splines are chosen as basis functions for all the applications. It is conjectured that B-splines bring about nearly orthogonal basis functions \citep{berry2002bayesian} and exhibits compact support property \citep{de1978practical}, i.e., non-zero only in short subinterval. The multidimensional B-splines are generated from the tensor product of the individual one-dimensional B-splines \citep{de1978practical}.

The PDE model in equation \ref{EQ01} is represented by the same linear combination of basis functions as

\begin{equation} \label{EQ04} 
	\mathcal{F}\left(\mathbf{x},{\mathbf{b}}^T\left(\mathbf{x}\right)\boldsymbol{\beta },\dots ,\{{\mathbf{b}}^T\left(\mathbf{x}\right)\boldsymbol{\beta }{\}}^q,\{\partial \mathbf{b}\left(\mathbf{x}\right)/\partial x_1{\}}^T\boldsymbol{\beta },\dots ;\boldsymbol{\theta }\right)=0,\qquad \mathbf{x}\in \boldsymbol{\varOmega} 
\end{equation} 

Instead of directly estimating the PDE parameters $\boldsymbol{\theta }=\left({\theta }_1,\dots ,{\theta }_m\right)$, the local parameters of the basis functions, $\boldsymbol{\beta }={\left({\beta }_1,\dots ,{\beta }_K\right)}^T$, are estimated from the noisy data by imposing the constraint that the data satisfies the underlying governing PDE $\mathcal{F}=0$ given in equation \ref{EQ01} for each of the observations.

Thus, the method of \textbf{SNAPE} solves the following constrained optimization problem:

\begin{equation} \label{EQ05} 
	\begin{array}{c}
		{\mathop{\mathrm{min}}_{\boldsymbol{\beta },\boldsymbol{\theta }}\ \ \ \  \sum^n_{i=1}{{\left\{y_i-{\mathbf{b}}^T\left({\mathbf{x}}_i\right)\boldsymbol{\beta }\right\}}^2}\ } \\ 
		subject\ to\ \ \ \ \mathcal{F}\left(\mathbf{x},{\mathbf{b}}^{\mathbf{T}}\left(\mathbf{x}\right)\boldsymbol{\beta },\dots ,\{{\mathbf{b}}^{\mathbf{T}}\left(\mathbf{x}\right)\boldsymbol{\beta }{\}}^q,\{\partial \mathbf{b}\left(\mathbf{x}\right)/\partial x_1{\}}^T\boldsymbol{\beta },\dots ;\boldsymbol{\theta }\right)=\mathbf{0},\qquad \mathbf{x}\mathrm{\in }\boldsymbol{\varOmega} \end{array}
\end{equation}	
	
Figure \ref{fig:1} illustrates the details of the proposed method of \textbf{SNAPE }for estimating the parameters of the PDE model using simultaneous basis function approximation. Even though the domain of the illustrated process in figure \ref{fig:1} is restricted to two dimensions for the purpose of visualization, the applicability of \textbf{SNAPE} can be generalized for any multidimensional PDE model.
    
\subsection{Wave equation in two space dimensions}

\begin{figure}
	\centering
	\includegraphics[width=1.0\textwidth]{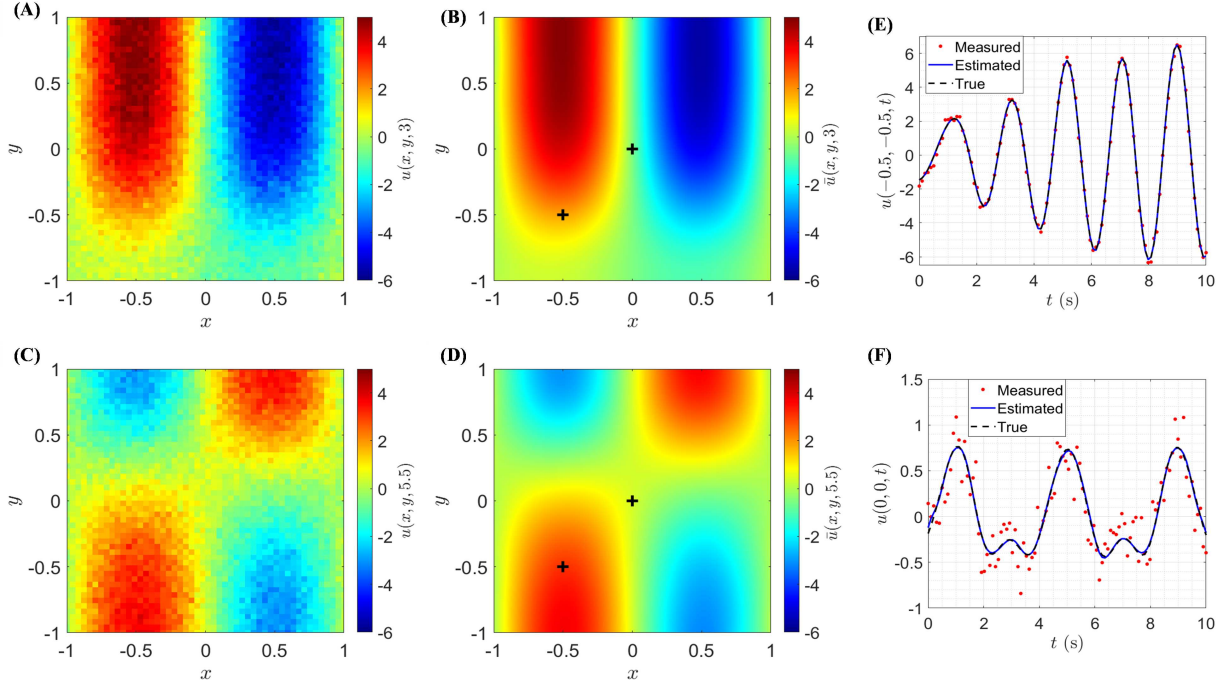}
	\caption{\textbf{Spatiotemporal response of the 2D wave equation.} The full-field measured response of the 2D wave PDE model at time instants of \textbf{(A) }$t=3$\textbf{ (C)}$t=5.5$ for an instance of 10\% Gaussian noise. The corresponding snapshots at \textbf{(B)}$t=3$ and \textbf{(D)}$t=5.5$ displays the smooth analytical approximation to the PDE solution estimated using \textbf{SNAPE}. The black plus markers denote the positions whose time histories are shown in \textbf{(E)} for $(x=-0.5,\ y=-0.5)$  and \textbf{(F)} for $(x=0,\ y=0)$. Even in the presence of moderate noise, \textbf{SNAPE} successfully approximates the true solution.}\label{fig:2}
\end{figure}

The wave equation represents PDE of the scalar function $u\left(\mathbf{x}\right)$ where the domain $\mathbf{x}\mathrm{\in }\left(x_1,x_2,\dots ,x_m;t\right)$ consists of a time variable and $m$ spatial variables. The PDE is expressed as ${\mathrm{u}}_{\mathrm{tt}}=c^2{\mathrm{\nabla }}^2u$ where $c$ is a real coefficient and ${\mathrm{\nabla }}^2$ is the Laplacian operator. This second-order linear PDE forms the basis of various fields of physics such as classical mechanics, quantum mechanics, geophysics, general relativity to name a few. The parameters of the PDE model bear information regarding the physical property of the medium through which the wave is propagating along the corresponding spatial direction. It is assumed that dense measurements of the dependent scalar quantity, which may be the pressure in a fluid medium or the displacement along a specific direction, are acquired using sensors. The goal of the present study is to infer the physics from the measured data. As the physics of the dynamic process is known to us which is expressed in the mathematical form of the PDE, we need to estimate its parameters to infer the properties of the media.

As an example, the numerical solution to the following PDE with parameters $\boldsymbol{\theta }=\left(1.0,1.0\right)$ is obtained which represents 2D wave propagation.

\begin{equation} \label{EQ06} 
	\frac{{\partial }^2u}{\partial t^2}={\theta }_1\frac{{\partial }^2u}{\partial x^2}+{\theta }_2\frac{{\partial }^2u}{\partial y^2} 
\end{equation} 

A square spatial dimension is selected with geometry $(x,y)\mathrm{\in }\left[-1.0,1.0\right]$ and time span of $t\mathrm{\in }\left[0,10\right]$. Both the Dirichlet $\left(u=0\right)$ and the Neumann $\left(\partial u/\partial y=0\right)$ boundary conditions are applied at the opposite edges of$\ x=-1.0$, $x=1.0,$ and $y=-1.0$, $y=1.0$ respectively. The initial condition of the dynamic process is set to $u\left(x,y,0\right)=3sin\left(\mathrm{\pi }x\right)exp\left(sin\left(\frac{\mathrm{\pi }}{2}y\right)\right)$ and $u_t\left(x,y,0\right)=tan^{-1}\left(cos\left(\frac{\pi }{2}x\right)\right)$. The generated response $u\left(x,y,t\right)\mathrm{\in }{\mathbb{R}}^{50\times 50\times 100}$ is corrupted with 10\% Gaussian noise to simulate measurement noise from the sensors. The proposed method of \textbf{SNAPE} is adopted to infer the PDE parameters. The mean of the estimated parameters $\overline{\boldsymbol{\theta }}=\left(1.002,1.022\right)$ of the PDE model exhibits superior accuracy from the noise corrupted measured data. The robustness to noise is further demonstrated by computing the coefficient of variation (\textit{cov}) of the estimates to be as low as $cov\boldsymbol{(}\boldsymbol{\theta }\boldsymbol{)=}\left(0.07,\ 0.60\right)\%$. It also estimates the analytical approximation of the solution to the PDE model without the knowledge of the initial and boundary conditions which generated the acquired dynamic response. Figures \ref{fig:2}(A) and \ref{fig:2}(C) show the measured response of the system with one such random instance of Gaussian noise at the time instants of $t=3$ and $t=5.5$ respectively. The estimated approximate solution from the discrete measurements consists of a smooth continuous function as shown in Figures \ref{fig:2}(B) and \ref{fig:2}(D) for the same corresponding time instants. The time histories of two localized positions are shown in figures \ref{fig:2}(E) and \ref{fig:2}(F) that compares the measured response and the estimated function with the true response of the system. It is evident that the estimated function of the solution satisfactorily approximates the true response.

\subsection{Chaotic response of forced Duffing oscillator}

\begin{figure}
	\centering
	\includegraphics[width=0.8\textwidth]{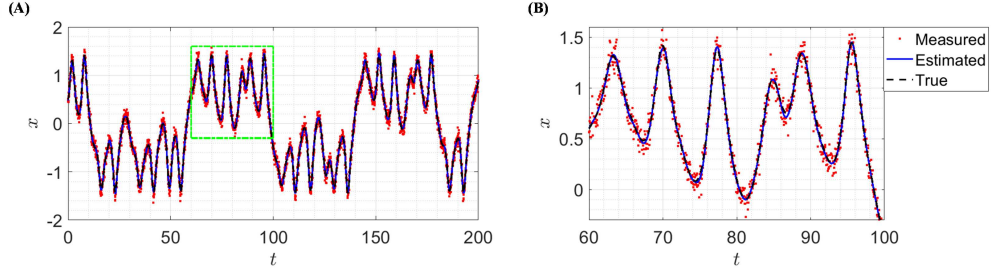}
	\caption{\textbf{Chaotic solution of forced Duffing equation.} The solution of the nonlinear ODE of forced Duffing oscillator exhibits deterministic chaos for certain values of parameters as discussed in the text. \textbf{(A)} One such instance of measured chaotic response with 10\% Gaussian noise. \textbf{(B) }The magnified time history demonstrates the ability of the proposed method to estimate the chaotic solution even from moderate noisy data.}\label{fig:3}
\end{figure}

The Duffing equation represents the nonlinear dynamics of a system with cubic nonlinearity. The parameters $\boldsymbol{\theta }=\left({\theta }_1,\ {\theta }_2,\ {\theta }_3\right)$ in the nonhomogeneous ODE $x_{tt}+{\theta }_1x_t+{\theta }_2x+{\theta }_3x^3=\gamma cos(\omega t)$ provides the linear damping and stiffness as well as the nonlinear cubic stiffness of the system. At the forcing parameters of $\gamma =0.42$ and $\omega \ =\ 1$ and system parameters of  $\boldsymbol{\theta }=\left(0.5,\ -1,\ 1\right)$ the solution of the nonlinear ODE exhibits deterministic chaos. For the provided values of the ODE parameters, the system is numerically solved for period $t\mathrm{\in }\left[0,\ 200\right]$ and the response $x(t)\mathrm{\in }{\mathbb{R}}^{4000}$ is perturbed with 10\% Gaussian noise to mimic measurement noise. One such random instance of measured data is compared with the true response in Figure \ref{fig:3}(A). Figure \ref{fig:3}(B) shows the magnified section of the small part of the data. \textbf{SNAPE} is applied to the noise corrupted chaotic response to infer the parameters of the system. The mean of the estimated parameters is $\overline{\boldsymbol{\theta }}=\left(0.49,\ -1.0,\ \ 0.99\right)$ and the corresponding uncertainty of estimation as$\ cov\boldsymbol{(}\boldsymbol{\theta }\boldsymbol{)=}\left(1.06,\ 0.98,\ 0.63\right)\%$ signifies the superior accuracy and robustness of the proposed method. Also, the analytical approximate solution of the Duffing equation compares well with the true solution as shown in Figures \ref{fig:3}(A) and \ref{fig:3}(B).

\subsection{Parameter estimation of Navier-Stokes equations}

\begin{figure}
	\centering
	\includegraphics[width=1.0\textwidth]{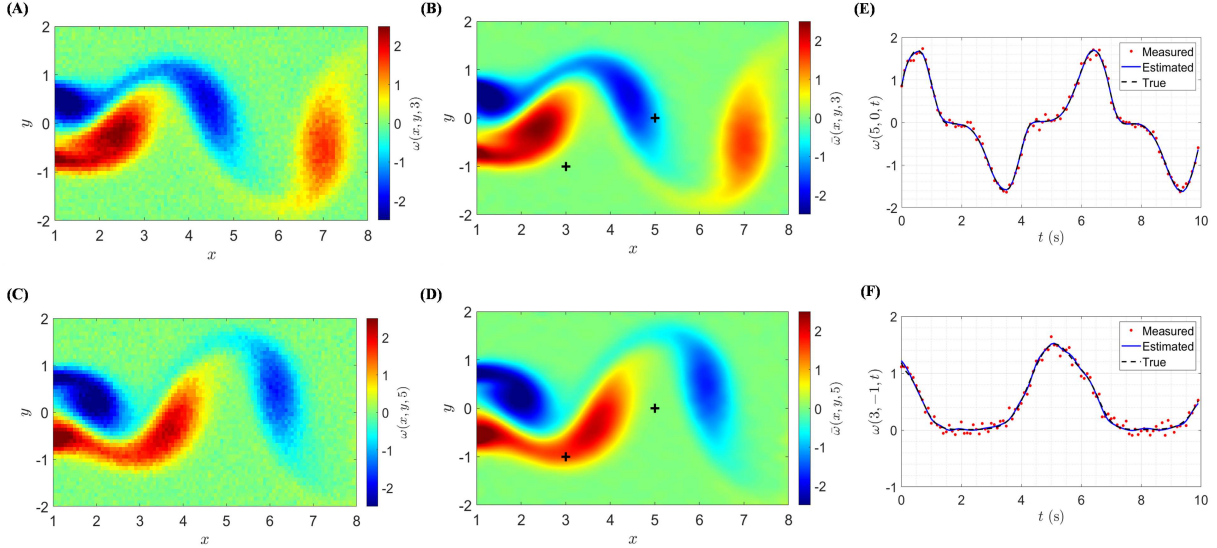}
	\caption{\textbf{Inferring Navier-Stokes equation from 10\% Gaussian added noise.} The full-field measured response of the Navier-Stokes PDE model shows vortex-shedding at time instants of \textbf{(A) }$t=3$\textbf{ (C)}$t=5$ for one random instance of 10\% Gaussian noise. \textbf{SNAPE }estimates smooth approximation to the solution from the noisy data whose corresponding snapshots at \textbf{(B)}$t=3$ and \textbf{(D)}$t=5$ are shown for comparison. The nonlinearity of the response is evident looking at the measured time histories of the positions \textbf{(E)} $(x=-0.5,\ y=-0.5)$  and \textbf{(F)} $(x=0,\ y=0)$. Regardless of the added noise and discretization error, \textbf{SNAPE} provides an estimated analytical solution of the PDE that satisfactorily approximates the hidden true solution.}\label{fig:4}
\end{figure}

The Navier-Stokes equations are a set of coupled nonlinear PDEs which describe the dynamics of fluids. The study of these equations is ubiquitous in a wide variety of scientific applications including climate modeling, blood flow in the human body, ocean currents, pollution analysis, and many more. This example involves incompressible flow past a cylinder which exhibits an asymmetric vortex shedding pattern in the wake of the cylinder. The equation in terms of the vorticity and velocity fields is given by

\begin{equation} \label{EQ07} 
	\frac{\partial \omega }{\partial t}+{\theta }_1\frac{{\partial }^2\omega }{\partial x^2}+{\theta }_2\frac{{\partial }^2\omega }{\partial y^2}+{\theta }_3u\frac{\partial \omega }{\partial x}+{\theta }_4v\frac{\partial \omega }{\partial y}=0 
\end{equation}

The two components of the velocity field data $u\left(x,y,t\right)$ and $v\left(x,y,t\right)$ are obtained from \citet{raissi2019physics} where the numerical solution of equation \ref{EQ07} is performed for the parameter values$\mathrm{\ }\boldsymbol{\theta }=\left(-0.01,-0.01,\ 1.0,1.0\right)$. The vorticity field data $\omega \left(x,y,t\right)$ is evaluated numerically from the velocity field data. The vorticity as well as the two components of velocity field datasets $\left(\omega ,u,v\right)\in {\mathbb{R}}^{100\times 50\times 100}$ are perturbed with 10\% Gaussian noise to simulate the measured data. The discrete measurement data is acquired over a rectangular domain of $x\mathrm{\in }\left[1.0,\ 8.0\right]$ and $y\mathrm{\in }\left[-2.0,\ 2.0\right]$ with the period of $t\mathrm{\in }\left[0,\ 9.9\right]$. The mean of the estimated parameters $\overline{\boldsymbol{\theta }}=\left(-0.01,-0.006,\ 0.88,\ 0.91\right)\ $with the uncertainty $cov\boldsymbol{(}\boldsymbol{\theta }\boldsymbol{)=}\left(5.46,\ 5.28,\ 5.33,\ 4.44\right)\%$ using the method of \textbf{SNAPE }compares satisfactorily well with the exact values considering the discretization error while evaluating vorticity from the velocity components.  Figures \ref{fig:4}(A) and \ref{fig:4}(C) show one instance of measured noise-corrupted vorticity field at $t=3$ and $t=5$ respectively. The corresponding smooth analytical approximation of the solution is shown in figures \ref{fig:4}(B) and \ref{fig:4}(C). The comparison of time histories of the estimated solution with the true response, at two different locations as shown in figures \ref{fig:4}(E) and \ref{fig:4}(F), corroborate the efficacy of the present method.

\subsection{Application in classical and quantum mechanics}

\begin{figure}
	\centering
	\includegraphics[width=1.0\textwidth]{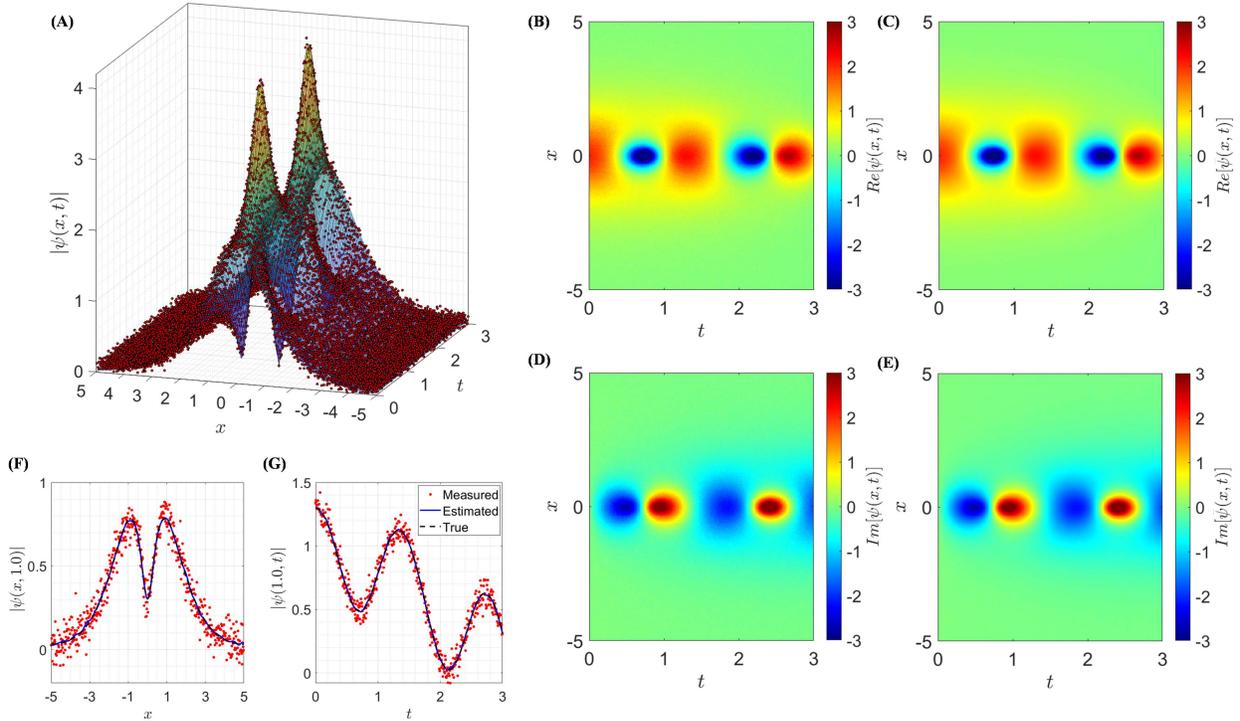}
	\caption{\textbf{Learning the nonlinear Schr\"{o}dinger equation from the complex field data.} \textbf{(A)} The magnitude of the complex field data $|\psi \left(x,t\right)|$ perturbed with a random realization of 10\% Gaussian noise overlaid on the surface of the true solution. \textbf{(B) }The real component of the measured complex field. \textbf{(C) }The real component of the estimated approximate solution. \textbf{(D)} The imaginary component of the measured complex field. \textbf{(E) }The imaginary component of the estimated approximate solution. The comparison of the magnitude of the estimated solution from the noisy complex field data to the magnitude of the true solution \textbf{(F) }at time instant $t=1$ and \textbf{(G)} at position $x=1$ reveals the efficacy and robustness of \textbf{SNAPE}.}\label{fig:5}
\end{figure}

The nonlinear Schr\"{o}dinger equation (NLSE) finds its application in light propagation through nonlinear optical fibers, the study of Bose-Einstein condensates, and small amplitude surface gravity waves. This example extends the applicability of the proposed method for complex fields $\psi \left(x,t\right)$  whose PDE is given as

\begin{equation} \label{EQ08} 
	\frac{\partial \psi }{\partial t}+{\theta }_1\frac{{\partial }^2\psi }{\partial x^2}+{\theta }_2{\left|\psi \right|}^2\psi =0 
\end{equation}

The data $\psi \left(x,t\right)\in {\mathbb{C}}^{512\times 501}$\textit{ }is obtained from \citet{rudy2017data} where the above PDE is numerically solved for the parameter values $\boldsymbol{\theta }=\left(-0.5i,-1.0i\right)$. The solution domain consists of $x\mathrm{\in }\left[-5,5\right]$ and $t\mathrm{\in }\left[0,\mathrm{\pi }\right]$. Like before, 10\% Gaussian noise is added to mimic the measurement data acquired using sensors. \textbf{SNAPE} is applied to the complex field measurement data, and with the domain knowledge of the structure of the governing PDE the mean of the estimated parameters is  $\overline{\boldsymbol{\theta }}=\left(-0.44i,-0.96i\right)$ with a low uncertainty bound of $cov\boldsymbol{(}\boldsymbol{\theta }\boldsymbol{)=}\left(0.76,\ 0.31\right)\%$. Figure \ref{fig:5}(A) shows the magnitude of an instance of noise corrupted measured complex field data superimposed on the true solution of the NLSE of equation \ref{EQ08}. The real and the imaginary components of the measured complex field data are shown in figures \ref{fig:5}(B) and \ref{fig:5}(D) respectively. \textbf{SNAPE} not only infers the parameters of the NLSE but also is successful in estimating the analytical approximate solution of NLSE. Figures \ref{fig:5}(C) and \ref{fig:5}(E) show the real and imaginary components of the estimated approximate solution. The efficacy of the proposed method is further exemplified in figures \ref{fig:5}(F) and \ref{fig:5}(G) where the magnitude of the analytical approximate solution estimated from noisy measured data is compared with the magnitude of the true solution at the time instant $t=1$  and the location $x=1$ respectively.

\subsection{Theory-guided learning of parametric ODEs and PDEs}

Table \ref{tab:01} exhibits the application of \textbf{SNAPE} on the measured response of a broad range of differential equation models predominant in the scientific community. The response includes both periodic as well as chaotic oscillations from one-dimensional time histories (ODEs) to multidimensional spatiotemporal dynamics (PDEs). The measured responses of all the systems reveal strong nonlinearity apart from the linear wave equation. For each of the models, the constrained equation in the optimization of Eq. 5 is custom-built following the convention of \textit{theory-guided learning}. The simulated real, as well as the complex field data, is corrupted with Gaussian noise to take into consideration the eminent noise from the sensors and acquisition devices. The robustness and repeatability of \textbf{SNAPE} are demonstrated by performing repeated estimation on 10 bootstrap samples of noise corrupted data. Unlike deep learning-based methods \citep{both2021deepmod}, \textbf{SNAPE} successfully learns the differential equations for each random instance of noisy data. Moreover, it provides uncertainty bounds of the estimated parameters that arise from the inherent randomness of the measurement noise and discretization errors. As the data for the PDEs of Kuramoto-Sivashinsky, Burgers', Korteweg-de Vries, and Schr\"{o}dinger equation are obtained from \citet{rudy2017data}, the results of the estimation provide a direct comparison of the regression-based method (\textit{9}) with the proposed method of \textbf{SNAPE}. For all four cases, the \textbf{SNAPE }exhibits higher accuracy and robustness to noise. The superior performance is more prominent in the case of higher-order PDEs like the Kuramoto-Sivashinsky equation where the accuracy of estimation of \textbf{SNAPE} on 5\% noise is much higher than that of the method in \citet{rudy2017data} on 1\% noise. The velocity field data of the Navier-Stokes equation is obtained from \citet{raissi2019physics} while the vorticity field data is computed from the velocity field data through numerical differentiation. Even though both the components of velocity and the vorticity data are corrupted with noise, the accuracy of the \textbf{SNAPE} estimates is similar to that of the deep learning-based method in \citet{raissi2019physics} for 1\% noise. Besides, \textbf{SNAPE }is successful in providing stable and robust estimates of the Navier-Stokes PDE parameters even for the higher amount of added noise. The results of the tabulated examples demonstrate the applicability and reliability of the proposed method for a wide variety of spatiotemporal processes where scientific theories are available.
\begin{landscape}
	\begin{table}[htbp]
		\centering
		\caption{\textbf{Parameter estimation of differential equation models prevalent in mathematical sciences. }For each of the examples, the standard form of the differential equations is provided along with the exact values of parameters used to simulate the responses. \textbf{SNAPE} is applied on 10 bootstrap samples generated from 1\% and 5\% Gaussian noise corrupted response for each of the examples. The mean $\overline{\boldsymbol{\theta }}$ and the coefficient of variation $cov\left(\boldsymbol{\theta }\right)$ of the estimated parameters demonstrates the accuracy and robustness of the proposed method.}
		\resizebox{\textwidth}{!}{%
			\begin{tabular}{|P{2.0in}|P{2.0in}|P{2.0in}|P{2.5in}|P{2.5in}|} \hline 
				{} & {} & {} & {} & {} \\[5pt] 
				\textbf{Differential Equations} & \textbf{Form} & \textbf{Exact} & \textbf{1\% Noise} & \textbf{5\% Noise} \\[10pt] \hline 
				{} & {} & {} & {} & {} \\[5pt]
				\textbf{Van der Pol oscillator} & $x_{tt}+{\theta }_1x_t+{\theta }_2x^2x_t+{\theta }_3x=0$ & $\boldsymbol{\theta }\boldsymbol{=}\left(-8,\ 8,\ 1\right)$ & $ \begin{array}{c}
					\overline{\boldsymbol{\theta }}\boldsymbol{=}\left(-7.95,\ 8.03,\ 1.01\ \right) \\ 
					cov\left(\boldsymbol{\theta }\right)\boldsymbol{=}\left(0.19,\ 0.19,\ 0.24\right)\% \end{array}
				$  & $ \begin{array}{c}
					\overline{\boldsymbol{\theta }}\boldsymbol{=}\left(-8.00,\ 8.08,\ 1.03\ \right) \\ 
					cov\left(\boldsymbol{\theta }\right)\boldsymbol{=}\left(1.84,\ 1.89,\ 1.30\right)\% \end{array}
				$   \\[20pt] \hline 
				{} & {} & {} & {} & {} \\[5pt]
				\textbf{Forced Duffing oscillator} & $x_{tt}+{\theta }_1x_t+{\theta }_2x+{\theta }_3x^3=0.42cos(t)$ & $\boldsymbol{\theta }\boldsymbol{=}\left(0.5,\ -1,\ 1\right)$ & $ \begin{array}{c}
					\overline{\boldsymbol{\theta }}\boldsymbol{=}\left(0.5,\ -0.99,\ 1.0\ \right) \\ 
					cov\boldsymbol{(}\boldsymbol{\theta }\boldsymbol{)=}\left(0.96,\ 0.89,\ 0.57\right)\% \end{array}
				$  & $ \begin{array}{c}
					\overline{\boldsymbol{\theta }}\boldsymbol{=}\left(0.49,\ -0.99,\ 1.0\ \right) \\ 
					cov\boldsymbol{(}\boldsymbol{\theta }\boldsymbol{)=}\left(1.0,\ 0.93,\ 0.60\right)\% \end{array}
				$  \\[20pt] \hline 
				{} & {} & {} & {} & {} \\[5pt]
				\textbf{2D Wave equation} & $u_{tt}={\theta }_1u_{xx}+{\theta }_2u_{yy}$\textit{ } & $\boldsymbol{\theta }\boldsymbol{=}\left(1,\ 1\right)$ & $ \begin{array}{c}
					\overline{\boldsymbol{\theta }}\boldsymbol{=}\left(1.00,\ 1.00\right) \\ 
					cov\boldsymbol{(}\boldsymbol{\theta }\boldsymbol{)=}\left(0.02,\ 0.18\right)\% \end{array}
				$  & $ \begin{array}{c}
					\overline{\boldsymbol{\theta }}\boldsymbol{=}\left(0.99,\ 1.02\right) \\ 
					cov\boldsymbol{(}\boldsymbol{\theta }\boldsymbol{)=}\left(0.02,\ 0.16\right)\% \end{array}
				$  \\[20pt] \hline 
				{} & {} & {} & {} & {} \\[5pt]
				\textbf{Kuramoto-Sivashinsky equation} & $u_t+{\theta }_1uu_x+{\theta }_2u_{xx}+{\theta }_3u_{xxxx}=0$  & $\boldsymbol{\theta }\boldsymbol{=}\left(1,\ 1,\ 1\right)$ & $ \begin{array}{c}
					\overline{\boldsymbol{\theta }}\boldsymbol{=}\left(1.06,\ 1.01,\ 1.01\right) \\ 
					cov\boldsymbol{(}\boldsymbol{\theta }\boldsymbol{)=}\left(0.89,\ 0.95,\ 0.93\right)\% \end{array}
				$\textbf{\textit{ \newline }}  & $ \begin{array}{c}
					\overline{\boldsymbol{\theta }}\boldsymbol{=}\left(0.88,\ 0.76,\ 0.76\right) \\ 
					cov\boldsymbol{(}\boldsymbol{\theta }\boldsymbol{)=}\left(21.8,\ 17.3,\ 17.9\right)\% \end{array}
				$  \\[20pt] \hline 
				{} & {} & {} & {} & {} \\[5pt]
				\textbf{Burgers' equation} & $u_t+{\theta }_1uu_x+{\theta }_2u_{xx}=0$\textit{ } & $\boldsymbol{\theta }\boldsymbol{=}\left(1,\ -0.1\right)$ & $ \begin{array}{c}
					\overline{\boldsymbol{\theta }}\boldsymbol{=}\left(1.01,\ -0.10\right) \\ 
					cov\boldsymbol{(}\boldsymbol{\theta }\boldsymbol{)=}\left(0.05,\ 0.11\right)\% \end{array}
				$  & $ \begin{array}{c}
					\overline{\boldsymbol{\theta }}\boldsymbol{=}\left(1.01,\ -0.10\right) \\ 
					cov\boldsymbol{(}\boldsymbol{\theta }\boldsymbol{)=}\left(0.17,\ 0.93\right)\% \end{array}
				$  \\[20pt] \hline 
				{} & {} & {} & {} & {} \\[5pt]
				\textbf{Korteweg-de Vries equation} & $u_t+{\theta }_1uu_x+{\theta }_2u_{xxx}=0$  & $\boldsymbol{\theta }\boldsymbol{=}\left(6,\ 1\right)$ & $ \begin{array}{c}
					\overline{\boldsymbol{\theta }}\boldsymbol{=}\left(6.02,\ 1.01\right) \\ 
					cov\boldsymbol{(}\boldsymbol{\theta }\boldsymbol{)=}\left(0.04,\ 0.08\right)\% \end{array}
				$  & $ \begin{array}{c}
					\overline{\boldsymbol{\theta }}\boldsymbol{=}\left(6.03,\ 1.03\right) \\ 
					cov\boldsymbol{(}\boldsymbol{\theta }\boldsymbol{)=}\left(0.18,\ 0.38\right)\% \end{array}
				$  \\[20pt] \hline 
				{} & {} & {} & {} & {} \\[5pt]
				\textbf{Nonlinear Schr\"{o}dinger equation} & ${\psi }_t+{\theta }_1{\psi }_{xx}+{\theta }_2{\left|\psi \right|}^2\psi =0$\textit{ } & $\boldsymbol{\theta }\boldsymbol{=}\left(-0.5i,\ -1i\right)$ & $ \begin{array}{c}
					\overline{\boldsymbol{\theta }}\boldsymbol{=}\left(-0.49i,\ -1.0i\right) \\ 
					cov\boldsymbol{(}\boldsymbol{\theta }\boldsymbol{)=}\left(0.05,\ 0.2\right)\% \end{array}
				$  & $ \begin{array}{c}
					\overline{\boldsymbol{\theta }}\boldsymbol{=}\left(-0.45i,\ -0.96i\right) \\ 
					cov\boldsymbol{(}\boldsymbol{\theta }\boldsymbol{)=}\left(0.44,\ 0.17\right)\% \end{array}
				$  \\[20pt] \hline 
				{} & {} & {} & {} & {} \\[5pt]
				\textbf{Navier-Stokes equation} & ${\omega }_t+{\theta }_1{\omega }_{xx}+{\theta }_2{\omega }_{yy}+{\theta }_3u{\omega }_x+{\theta }_4v{\omega }_y=0$ & $\boldsymbol{\theta }\boldsymbol{=}\left(-0.01,\ -0.01,\ 1,\ 1\right)$ & $ \begin{array}{c}
					\overline{\boldsymbol{\theta }}\boldsymbol{=}\left(-0.01,\ -0.01,\ 1.01,\ 1.02\right) \\ 
					cov\boldsymbol{(}\boldsymbol{\theta }\boldsymbol{)=}\left(0.02,\ 0.14,\ 0.06,\ 0.05\right)\% \end{array}
				$  & $ \begin{array}{c}
					\overline{\boldsymbol{\theta }}\boldsymbol{=}\left(-0.01,\ -0.01,\ 0.98,\ 0.99\right) \\ 
					cov\boldsymbol{(}\boldsymbol{\theta }\boldsymbol{)=}\left(1.90,\ 1.82,\ 1.39,\ 1.13\right)\% \end{array}
				$  \\[20pt] \hline 
		\end{tabular}}
		\label{tab:01}%
	\end{table}%
\end{landscape}

\subsection{Robustness to extreme noise}

\begin{figure}
	\centering
	\includegraphics[width=1.0\textwidth]{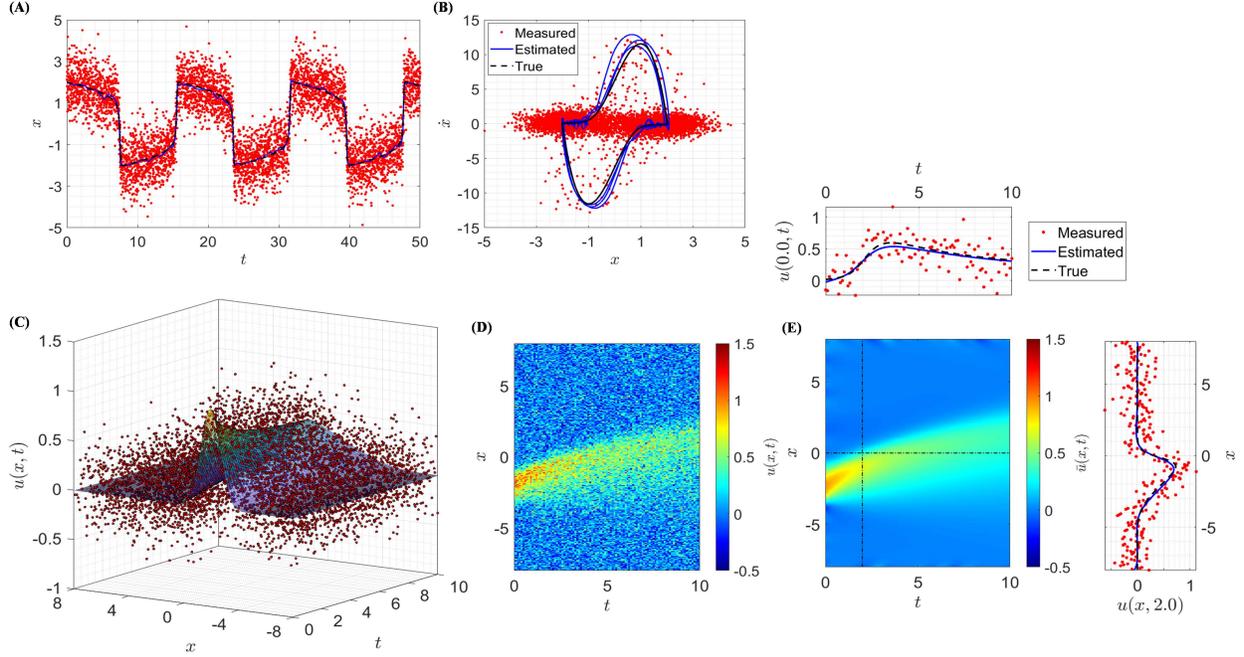}
	\caption{\textbf{Performance of SNAPE under extreme noise.} \textbf{(A)} The estimated functional solution approximates the true solution of the Van der Pol equation exhibiting nonlinear relaxation oscillation using \textbf{SNAPE} from measured time history perturbed with 50\% Gaussian noise. \textbf{(B) SNAPE} reveals the dominant phase portrait hidden within the cluster of noisy data. \textbf{(C) }The cloud of 100\% Gaussian noise corrupted response of Burgers' PDE model overlaid on its true response surface. \textbf{(D)} The measured response due to the presence of extreme noise vaguely acquires the nonlinear traveling wave. \textbf{(E) }Even in the presence of such extreme noise, \textbf{SNAPE} not only estimates the parameters of the PDE model with reasonable accuracy, it also estimates the analytical solution that satisfactorily approximates the true solution as revealed from the cross-sections of the response corresponding to the dotted lines.}\label{fig:6}
\end{figure}

In this part, an attempt is made to infer PDE model parameters and estimate its approximate solution using \textbf{SNAPE} from measured data having extreme levels of noise. In practice, there are situations where an acquired signal contains elevated noise due to the specified limitations of the sensor or acquisition system. Often, we tend to discard those measurements as it is difficult to infer useful information regarding the physical properties of those processes that govern the acquired response. In such scenarios, we can apply the scientific domain knowledge we have about the process and try to infer as much physics from the extremely noisy data as possible. \textbf{SNAPE} bridges the gap between well-established scientific theories and the latest data-driven learning algorithms.

The first example consists of the Van der Pol oscillator which exhibits non-conservative relaxation oscillations with nonlinear damping. Such relaxation oscillations are used in diverse physical and biological sciences, including but not limited to nonlinear electric circuits, geothermal geysers, networks of firing nerve cells, and the beating of the human heart. The evolution in time of the position $x$ is expressed by the differential equation $x_{tt}-\mu \left(1-x^2\right)x_t+x=0$\textit{ } where $\mu $ is the nonlinear parameter that regulates the strength of damping and relaxation. In a more general form, the following ODE model is used to generate the data.

\begin{equation} \label{EQ09} 
	\frac{d^2x}{dt^2}+{\theta }_1\frac{dx}{dt}+{\theta }_2x^2\frac{dx}{dt}+{\theta }_3x=0 
\end{equation}

The generated time history $x\left(t\right)\mathrm{\in }{\mathbb{R}}^{5000}$ for a period of $t\mathrm{\in }\left[0,50\right]$ with true parameter values of $\boldsymbol{\theta }=\left(-8.0,\ 8.0,\ 1.0\right)$ is corrupted with 50\% Gaussian noise to simulate extreme measurement noise. Even in the presence of acute noise in the measured signal as shown in figure \ref{fig:6}(A), the estimated solution function approximates well the true response of the system. Also, the mean of the parameters of the ODE $\overline{\boldsymbol{\theta }}=\left(-7.56,\ 7.94,\ 1.03\right)$ are estimated with reasonable accuracy. Even with such high noise content, the parameters are estimated with reasonable uncertainty of $cov\boldsymbol{(}\boldsymbol{\theta }\boldsymbol{)=}\left(29.4,\ 36.1,\ 13.67\right)\%$. As shown in figure \ref{fig:6}(B), the phase portrait of the measured response is too smudged to outline the hidden dynamics, whereas \textbf{SNAPE} approximately brings out the true phase portrait.

In the next example, the parameters of the Burgers' equation are estimated from its response which is perturbed with 100\% Gaussian noise. This nonlinear PDE occurs in many branches of applied mathematics such as fluid mechanics, gas dynamics, nonlinear acoustics, or traffic flows. The Burgers' equation is obtained from the Navier-Stokes equation by neglecting the term corresponding to the pressure gradient. Depending on the application, the parameters of the PDE model signify diffusion coefficient in gas dynamics or kinematic viscosity in fluid mechanics. The PDE model of the Burgers' equation is given as 

\begin{equation} \label{EQ10} 
	\frac{\partial u}{\partial t}+{\theta }_1u\frac{\partial u}{\partial x}+{\theta }_2\frac{{\partial }^2u}{\partial x^2}\ =\ 0 
\end{equation} 

The data $u\left(x,t\right)\mathrm{\in }{\mathbb{R}}^{256\times 101}$ is obtained from \citet{rudy2017data} for parameter values $\overline{\boldsymbol{\theta }}=\left(1.0,\ -0.1\right)$ with solution domain $x\mathrm{\in }\left[-8,\ 8\right]$ and $t\mathrm{\in }\left[0,\ 10\right]$. Figure \ref{fig:6}(C) shows the cloud of measurement data which is indistinguishable from the superimposed true response. \textbf{SNAPE} is applied to this extremely noisy data, with the knowledge of the mathematical form of the underlying process. The mean of the estimated parameters $\overline{\boldsymbol{\theta }}=\left(1.15,-0.19\right)$ demonstrates compromised accuracy due to such extreme noise content, yet the inference of the proposed estimation method is successful with the estimated uncertainty about the mean as $cov\boldsymbol{(}\boldsymbol{\theta }\boldsymbol{)=}\left(1.87,\ 6.61\right)\%$. Figure \ref{fig:6}(E) shows the approximate functional solution along with the cross-section of the responses at specific locations and instant of time.

\section{Discussion}
\label{sec:Discussion}

\begin{figure}
	\centering
	\includegraphics[width=0.4\textwidth]{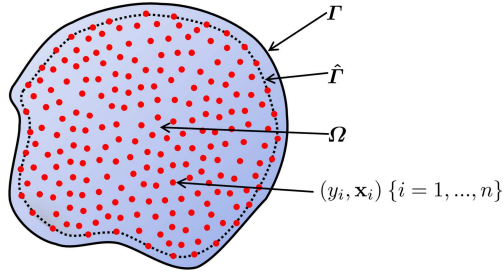}
	\caption{\textbf{Domain and boundary of a hypothetical differential equation.} A representative two-dimensional domain $\boldsymbol{\mathit{\Omega}}$ and the boundary $\boldsymbol{\mathit{\Gamma}}$ of an arbitrary PDE model. The red dots indicate $n$ number of discrete measurements $\left(y_i,{\mathbf{x}}_i\right)$. The dotted curve $\widehat{\boldsymbol{\mathit{\Gamma}}}$ represents pseudo-boundary of the PDE model defined by the peripheral data points located on it.}\label{fig:7}
\end{figure}

\textbf{SNAPE} explicitly satisfies the differential equation $\mathcal{F}=0$ in the form of constraints in the optimization, however, it does not require the knowledge of the initial or the boundary conditions. As per the formulation of the optimization problem of \textbf{SNAPE}, the initial, as well as the boundary conditions, are implicitly satisfied at $\mathbf{x}\boldsymbol{\in }\widehat{\boldsymbol{\mathit{\Gamma}}}$, a sub-domain of $\boldsymbol{\varOmega}$ as shown in figure \ref{fig:7}. The measurement points at the periphery of the domain $\boldsymbol{\varOmega}$ form a pseudo-boundary $\widehat{\boldsymbol{\mathit{\Gamma}}}$ represented by the dotted closed curve in figure 7. By minimizing the loss function of \textbf{SNAPE} in Eq. 5, the Dirichlet boundary condition of $g\left({\mathbf{x}}_i\right)\approx y_i\ $is approximately satisfied where $\left[\left(y_i,{\mathbf{x}}_i\right)\mathrm{\in }\widehat{\boldsymbol{\varGamma}}\right]$. This implies \textbf{SNAPE }can learn the PDE models from the data acquired from inside the domain irrespective of the initial or the boundary conditions. The learned differential equation (ODE and PDE) models enable us to simulate responses for initial or boundary conditions other than that of the observed response. Besides estimating the parameters of the model, \textbf{SNAPE }provides an analytical approximation for the solution of the differential equation $g\left(\mathbf{x}\right)\approx \overline{g}\left(\mathbf{x}\right)={\mathbf{b}}^T\left(\mathbf{x}\right)\boldsymbol{\beta }$. It signifies that the approximate response of the governing process can be evaluated from the continuous function $\overline{g}\left(\mathbf{x}\right)$ for any real value of  $\mathbf{x}\boldsymbol{\in }\boldsymbol{\varOmega}$,even though the response is observed at discrete points. Furthermore, \textbf{SNAPE} avoids the evaluation of numerical derivatives that sets it apart from other regression-based methods. As a result, it provides a stable estimation of the model parameters even from responses with high noise content. Compared to the deep learning-based methods, \textbf{SNAPE} demonstrates higher robustness and repeatability in the learning of the model as the estimation is performed with 10 random bootstrap realizations of noise corrupted responses for all the applications.

Unlike data-driven machine learning techniques, the indispensable component of \textbf{SNAPE} is the known theory of the dynamic process that is derived from the first principle. It combines the domain knowledge that we have studied and discovered so far with the modern aspects of data science to infer the differential equation models from the observed data. This theory-specific subjectivity of the estimation framework is attributed to the formulation of the constrained equation in \textbf{SNAPE} for each application. In situations where two or more theories are hypothesized for a set of observed data, \textbf{SNAPE} can be extended to include the competing classes of differential equations in its optimization scheme to perform model selection. In the current version, \textbf{SNAPE} enforces an ODE or a PDE as a constraint, a future extension will be the incorporation of coupled ODEs or PDEs into the optimization scheme so that it can simultaneously estimate parameters of the system of differential equations. Even though Table S2 in supplementary materials compares the performance of \textbf{SNAPE} with that of the deep learning-based method for the Navier-Stokes equation, the future scope of work will include a more comprehensive comparison of their respective benefits and limitations for wider applications. \textbf{SNAPE} can be used to address the much-unexplored theory of identifiability of nonlinear differential equation models from a set of observations.  This in turn will not only enrich our understanding of nonlinear differential equations (ODEs and PDEs) but also promote smart strategies of nonlinear control and sensor placement for complex dynamic processes. 

\section{Materials and Methods}
\label{sec:Methods}

The proposed \textbf{SNAPE }algorithm performs the constrained optimization of equation \ref{EQ05} by searching for the optimal $\overline{\boldsymbol{\beta }}$ that minimizes the loss function and simultaneously satisfies the constrained equation parameterized by $\overline{\boldsymbol{\theta }}$ that approximates the governing differential equations. \textbf{SNAPE} is performing the task of inferring the parameters of the differential equations by avoiding the computation of the higher-order derivatives and subsequently avoids infusion of unnecessary numerical errors in the process of estimation.

\subsection{Formulation of the optimization problem}

The form of the constrain equation depends on the form of the underlying differential equation, so the exact algorithm of \textbf{SNAPE} slightly varies with each model yet the framework of estimation remains the same. For example, the shorthand notation of the functional relation that approximates the Burgers' equation \ref{EQ10} is given as.

\begin{equation} \label{EQ11} 
	\mathcal{F}\left(\mathbf{x},{\mathbf{b}}^T\left(x,t\right)\boldsymbol{\beta },{\left(\frac{\mathrm{\partial }\mathbf{b}\left(x,t\right)}{\mathrm{\partial }t}\right)}^T\boldsymbol{\beta },{\left(\frac{\mathrm{\partial }\mathbf{b}\left(x,t\right)}{\mathrm{\partial }x}\right)}^T\boldsymbol{\beta },{\left(\frac{{\mathrm{\partial }}^2\mathbf{b}\left(x,t\right)}{\mathrm{\partial }x^2}\right)}^T\boldsymbol{\beta };\boldsymbol{\theta }\right)\mathrm{\approx }\mathbf{0} 
\end{equation}

Now, the basis functions ${\mathbf{b}}^T\left(x,t\right)$ are evaluated at $n$ observation points to obtain basis matrix $\overline{\mathbf{B}}\mathrm{\in }{\mathbb{R}}^{n\mathrm{\times }m}$, where $m$ is the number of columns in the basis matrix which depends on the choice of the order and number of knots in the B-splines functions. The order of the B-spline basis functions ${\mathbf{b}}^T\left(x,t\right)$ are chosen such that it can be differentiated up to the degree of the PDE. Likewise, the following matrices are evaluated as well.

\begin{equation} \label{EQ12} 
	\begin{aligned}
	\frac{\mathrm{\partial }u}{\mathrm{\partial }t}\mathrm{\approx }{\left(\frac{\mathrm{\partial }\mathbf{b}\left(x,t\right)}{\mathrm{\partial }t}\right)}^T\boldsymbol{\beta }=&{\overline{\mathbf{B}}}_{\mathbf{0}}\boldsymbol{\beta } \\
	\frac{\partial u}{\partial x}\mathrm{\approx }{\left(\frac{\partial \mathbf{b}\left(x,t\right)}{\partial x}\right)}^T\boldsymbol{\beta }=&{\overline{\mathbf{B}}}_{\mathbf{1}}\boldsymbol{\beta } \\
	\frac{{\partial }^2u}{\partial x^2}\mathrm{\approx }{\left(\frac{{\partial }^2\mathbf{b}\left(x,t\right)}{\partial x^2}\right)}^T\boldsymbol{\beta }=&{\overline{\mathbf{B}}}_{\mathbf{2}}\boldsymbol{\beta }
	\end{aligned}
\end{equation}

where ${\overline{\mathbf{B}}}_{\mathbf{0}},{\overline{\mathbf{B}}}_{\mathbf{1}},{\overline{\mathbf{B}}}_{\mathbf{2}}\mathrm{\in }{\mathbb{R}}^{n\mathrm{\times }m}$ . The measured data is fitted with the B-spline functions such that at every point of measurement the PDE of equation \ref{EQ10} is satisfied, or the condition in equation \ref{EQ11} is satisfied. Hence, the optimization problem as presented in equation \ref{EQ05} is recast into the following form.

\begin{equation} \label{EQ13} 
	\begin{array}{c}
		{\mathop{\mathrm{min}}_{\boldsymbol{\beta },{\theta }_1,{\theta }_2}\ \ \ \  \frac{1}{2}||\mathbf{y}-\overline{\mathbf{B}}\boldsymbol{\beta }{||}^2_2\ } \\[5pt]
		subject\ to\ \ \ \ {\overline{\mathbf{B}}}_{\mathbf{0}}\boldsymbol{\beta }\boldsymbol{+}{\theta }_1\overline{\mathbf{B}}\boldsymbol{\beta }\boldsymbol{\bigodot }{\overline{\mathbf{B}}}_{\mathbf{1}}\boldsymbol{\beta }\boldsymbol{+}{\theta }_2{\overline{\mathbf{B}}}_{\mathbf{2}}\boldsymbol{\beta }\boldsymbol{\le }\boldsymbol{\delta } \end{array}	
\end{equation}

where $\mathrm{\odot }$ represents Hadamard (elementwise) product. Due to the presence of measurement noise $\boldsymbol{\epsilon }$ as well as discretization error, the residual of the approximate PDE model used in the constraint equation is not equated to zero but bounded by a small magnitude of modeling error $\boldsymbol{\delta }$.

\subsection{Alternating Direction Method of Multipliers (ADMM)}

\begin{figure}
	\centering
	\includegraphics[width=0.5\textwidth]{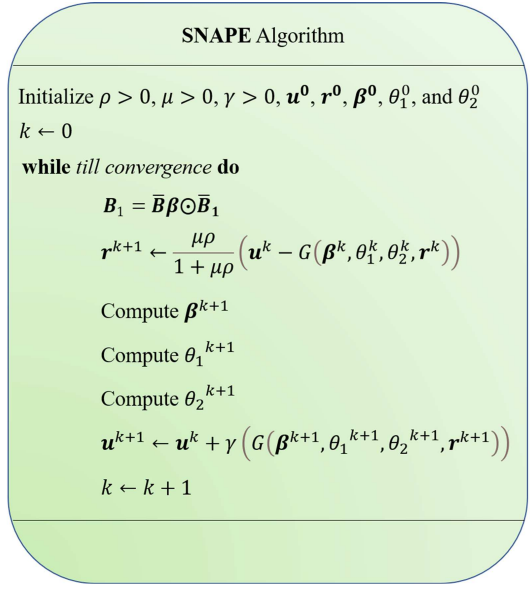}
	\caption{\textbf{SNAPE algorithm for Burgers' equation.} As per the notion of \textit{theory-guided learning}, the constraint equation in the optimization framework of \textbf{SNAPE} is unique for a model. Although the provided \textbf{SNAPE} algorithm is explicitly applicable for Burgers' PDE model, it demonstrates the key components of the algorithm which can be easily extended to any other linear or nonlinear models (both ODEs and PDEs).}\label{fig:8}
\end{figure}

This section describes the ADMM algorithm to solve the constrained optimization of the \textbf{SNAPE} method as stated in Eq. 13. The ADMM algorithm has originally been proposed by \citet{gabay1976dual} to find the infimum of variational problems that appear in continuum mechanics. The equivalent representation \citep{yang2011alternating} of the optimization problem in Eq. 13 is given as

\begin{equation} \label{EQ14} 
	\begin{array}{c}
		{\mathop{\mathrm{min}}_{\boldsymbol{\beta },{\theta }_1,{\theta }_2}\ \ \ \  \frac{1}{2}||\mathbf{y}-\overline{\mathbf{B}}\boldsymbol{\beta }{||}^2_2\ }+\frac{1}{2\mu }||\mathbf{r}{||}^2_2 \\[5pt] 
		subject\ to\ \ \ \ {\overline{\mathbf{B}}}_{\mathbf{0}}\boldsymbol{\beta }\boldsymbol{+}{\theta }_1\overline{\mathbf{B}}\boldsymbol{\beta }\boldsymbol{\bigodot }{\overline{\mathbf{B}}}_{\mathbf{1}}\boldsymbol{\beta }\boldsymbol{+}{\theta }_2{\overline{\mathbf{B}}}_{\mathbf{2}}\boldsymbol{\beta }\boldsymbol{+}\mathbf{r}\boldsymbol{=}\mathbf{0} \end{array}	
\end{equation} 

where $\mathbf{r}\mathrm{\in }{\mathbb{R}}^n$ is an auxiliary variable. The scaled form of augmented Lagrangian of the above optimization problem is given as

\begin{equation} \label{EQ15} 
	\mathcal{L}\left(\beta ,{\theta }_1,{\theta }_2,u,r\right)=\frac{1}{2}||\mathbf{y}-\overline{\mathbf{B}}\boldsymbol{\beta }{||}^2_2+\frac{1}{2\mu }||\mathbf{r}{||}^2_2+\frac{\rho }{2}||G\left(\beta ,{\theta }_1,{\theta }_2,r\right)+\mathbf{u}{||}^2_2-\frac{\rho }{2}||\mathbf{u}{||}^2_2 
\end{equation} 

where the function $G\left(\boldsymbol{\beta },{\theta }_1,{\theta }_2,\mathbf{r}\right)={\overline{\mathbf{B}}}_{\mathbf{0}}\boldsymbol{\beta }\boldsymbol{+}{\theta }_1\overline{\mathbf{B}}\boldsymbol{\beta }\boldsymbol{\bigodot }{\overline{\mathbf{B}}}_{\mathbf{1}}\boldsymbol{\beta }\boldsymbol{+}{\theta }_2{\overline{\mathbf{B}}}_{\mathbf{2}}\boldsymbol{\beta }\boldsymbol{+}\mathbf{r}$. The ADMM optimization \citep{boyd2011distributed} scheme involves an iterative update of the optimization parameters till its convergence. In the case of linear differential equation models, the function $G\left(\right)$ will be linear in terms of the basis coefficients $\boldsymbol{\beta }$, rendering the problem in equation \ref{EQ13} as biconvex optimization. It means in one of the iteration updates steps, the subproblem is convex with respect to one of the parameters by treating the other parameter as constant. In the case of nonlinear models such as here, the matrix$\boldsymbol{\mathrm{\ }}{\mathbf{B}}_{\mathbf{1}}\boldsymbol{\mathrm{=}}\overline{\mathbf{B}}\boldsymbol{\beta }\boldsymbol{\bigodot }{\overline{\mathbf{B}}}_{\mathbf{1}}$ is assumed constant for each iteration so that the function $G\left(\right)$ becomes linear in terms of $\boldsymbol{\beta }$. It is a biconvex relaxation of the original nonconvex problem when nonlinear differential equations are considered. The updates of the parameters at $k$${}^{th}$ step are computed by the following ADMM form \citep{yang2011alternating, boyd2011distributed}.

\begin{equation} \label{EQ16} 
	\begin{aligned}
		{\mathbf{r}}^{k+1}\coloneqq& \frac{\mathrm{\mu }\mathrm{\rho }}{1+\mathrm{\mu }\mathrm{\rho }}\left({\mathbf{u}}^k-G\left({\mathrm{\beta }}^k,{\theta }^k_1,{\theta }^k_2,{\mathbf{r}}^k\right)\right) \\
		{\boldsymbol{\beta }}^{k+1}\coloneqq& \ \ \mathop{\mathrm{argmin}}_{\boldsymbol{\beta }}\left(\mathcal{L}\left({\boldsymbol{\beta }}^k,{\theta }^k_1,{\theta }^k_2,\mathbf{u},{\mathbf{r}}^{k+1}\right)\right) \\
		{{\theta }_1}^{k+1}\coloneqq& \ \ \mathop{\mathrm{argmin}}_{{\theta }_1}\left(\mathcal{L}\left({\boldsymbol{\beta }}^{k+1},{\theta }^k_1,{\theta }^k_2,\mathbf{u},{\mathbf{r}}^{k+1}\right)\right) \\
		{{\theta }_2}^{k+1}\coloneqq& \ \ \mathop{\mathrm{argmin}}_{{\theta }_2}\left(\mathcal{L}\left({\boldsymbol{\beta }}^{k+1},{{\theta }_1}^{k+1},{\theta }^k_2,\mathbf{u},{\mathbf{r}}^{k+1}\right)\right) \\
		{\mathbf{u}}^{k+1}\coloneqq& {\mathbf{u}}^k+\gamma \left(G\left({\boldsymbol{\beta }}^{k+1},{{\theta }_1}^{k+1},{{\theta }_2}^{k+1},{\mathbf{r}}^{k+1}\right)\right)
	\end{aligned}
\end{equation}

The \textbf{SNAPE }algorithm for the Burgers' equation is provided in figure \ref{fig:8}. The updates of the parameters at each iteration step of the algorithm are computed by optimizing the corresponding objectives in Eq. 15. The closed-form expressions of the optimal parameters at each iteration step are obtained due to the aforementioned biconvex relaxation. For other ODEs or PDEs, a similar computational framework is followed by tweaking the provided algorithm with the corresponding form of the $G\left(\right)$ function.

\section*{Acknowledgment}
\label{S:ack}
The authors wish to acknowledge Dr. Anastasios Kyrillidis, assistant professor in the Department of Computer Science at Rice University for his valuable discussions on the ADMM optimization framework. This research was made possible by Science and Engineering Research Board of India (SERB)-Rice University Fellowship to Sutanu Bhowmick for pursuing his Ph.D. at Rice University. The financial support by SERB-India is gratefully acknowledged.

\renewcommand\thefigure{A.\arabic{figure}}
\setcounter{figure}{0}
\renewcommand{\theequation}{A.\arabic{equation}}
\setcounter{equation}{0}
\renewcommand\thetable{A.\arabic{table}}
\setcounter{table}{0}
\section*{Appendix}

This section provides detailed additional information regarding the proposed method of \textbf{SNAPE}. At first, the univariate B-spline basis function which forms the building block of \textbf{SNAPE} is discussed in brief along with its extension for multidimensional functions. Then the closed-form expression of the optimum parameters at each iterative ADMM update of the algorithm is derived. Further, the convergence of \textbf{SNAPE} for responses corrupted with various amounts of noise and random initialization is extensively studied. The examples of the Korteweg-de Vries equation and the Kuramoto-Sivashinsky equation that are included in Table 1, are discussed in detail in this supplementary document. Finally, the performance of \textbf{SNAPE} is compared with the previously proposed methods in the literature.

\subsection*{B-spline basis function}

\begin{figure}
	\centering
	\includegraphics[width=0.7\textwidth]{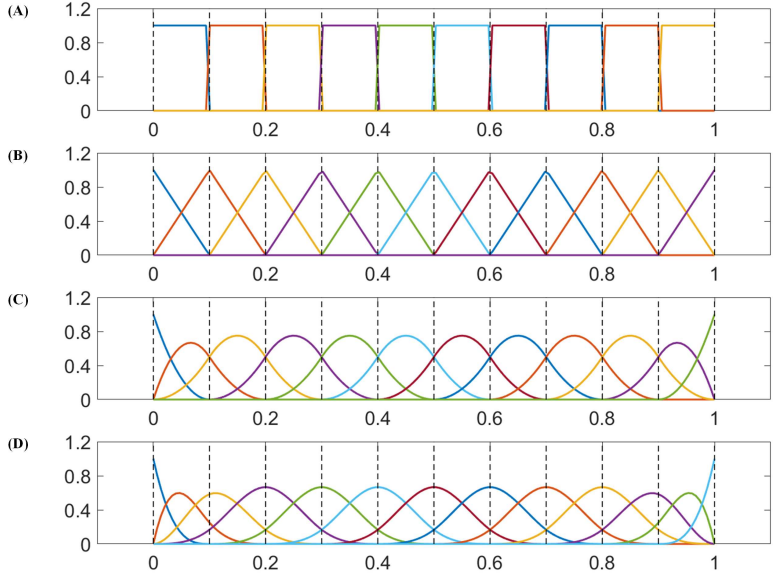}
	\caption{The sequence of B-spline basis functions of \textbf{(A)} order 1, \textbf{(B)} order 2, \textbf{(C)} order 3, and \textbf{(D)} order 4 with 11 knots evenly spaced between 0 and 1. Each B-spline basis function is non-zero on a few adjacent subintervals, hence they have local support.}\label{fig:S1}
\end{figure}

A univariate B-spline is a polynomial function of specific order defined over a domain with $k$ number of knots in equal or unequal intervals including the two boundaries. \citet{de1978practical} provides a recursive algorithm to generate B-splines of any order from B-splines of lower order. Figure \ref{fig:S1} shows a sequence of B-splines up to order four for the domain $[0,1]$ with 11 equidistant knots shown by the dashed vertical lines. The individual B-spline basis function is non-zero within a small interval, thereby demonstrating its property of compact (local) support. The number of basis functions with $k$ knots is computed as $p=k+o-2$  where $o$ is the order of the B-splines. The polynomial pieces join at $o$ inner knots where the derivatives up to orders $(o-1)$ are continuous. In the present study, the univariate B-spline basis functions are generated using the functional data analysis Matlab toolbox \citep{ramsay2002applied}.

The univariate B-spline basis functions are extended to obtain the multidimensional tensor product B-spline basis functions \citep{de1978practical, piegl1996nurbs, eilers2003multivariate}. For example, a two-dimensional domain $\mathbf{x}\mathrm{\in }{\mathbb{R}}^2\ $consisting of one spatial dimension and another temporal dimension $\mathbf{x}\mathrm{\in }\left(x,t\right)$ will have a set of basis functions ${\mathbf{b}}_{1p}\left(x\right),p=1,\dots ,m_1$ to represent functions in the $x$ domain, and similarly a set of $m_2$ basis functions ${\mathbf{b}}_{2p}\left(t\right),p=1,\dots ,m_2$ for the coordinate $t$. Then each of the $m_1\times m_2$ tensor product basis functions are defined as

\begin{equation} \label{SEQ01} 
	{\mathbf{b}}_{jk}\left(\mathbf{x}\right)={\mathbf{b}}_{1j}\left(x\right){\mathbf{b}}_{2k}\left(t\right),j=1,\dots ,m_1,k=1,\dots ,m_2 
\end{equation} 

The tensor product B-spline basis function existing in the $x\times t$ plane is represented by the following two-dimensional function

\begin{equation} \label{SEQ02} 
	\mathbf{b}\left(\mathbf{x}\right)=\sum^{m_1}_{j=1}{\sum^{m_2}_{k=1}{{\beta }_{jk}}}{\mathbf{b}}_{jk}\left(\mathbf{x}\right) 
\end{equation}

\begin{figure}
	\centering
	\includegraphics[width=0.8\textwidth]{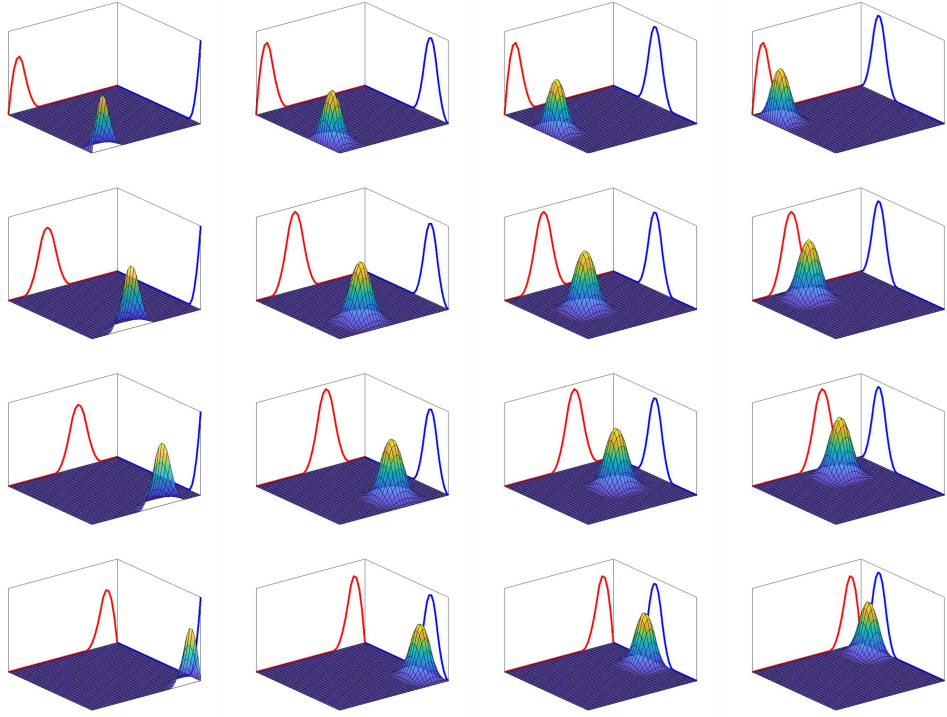}
	\caption{A portion of B-splines tensor product basis from some selected pairs of cubic B-splines. Each two-dimensional basis function is the tensor product of the corresponding one-dimensional B-spline basis functions.}\label{fig:S2}
\end{figure}

where ${\beta }_{jk}$ are the elements of $m_1\times m_2$ matrix of unknown tensor product B-spline coefficients. Figure S2 demonstrates 16 tensor product basis functions corresponding to the univariate cubic B-splines shown in blue and red, which is only a portion of a full-basis.  Each of the tensor product basis is positive corresponding to the nonzero support of the individual univariate ranges. The tensor product basis function of equation \ref{SEQ02} represents a continuous function that can be evaluated for any real value of the domain $\mathbf{x}$. The function $\mathbf{b}\left(\mathbf{x}\right)$ is evaluated at $n$ observation points within a grid of $n_x{\times n}_t$ in the $\mathbf{x}\mathrm{\in }\left(x,t\right)$ domain. The surface equation is re-expressed in matrix notation to incorporate computational efficiency as ${\mathbf{b}\left(\mathbf{x}\right)}_{n_x{\times n}_t}\boldsymbol{=}\mathbf{B}\ \boldsymbol{\beta }$ where $\boldsymbol{\beta }\boldsymbol{=}\boldsymbol{vec}\left(\left[{\beta }_{jk}\right]\right)$ and

\begin{equation} \label{SEQ03} 
	\mathbf{B}=\left({\mathbf{B}}_x\otimes {\mathbf{1}}^{\mathbf{T}}_{n_x}\right)\odot \left({\mathbf{1}}^{\mathbf{T}}_{n_t}\otimes {\mathbf{B}}_t\right) 
\end{equation} 

The matrices ${\mathbf{B}}_x\mathrm{\in }{\mathbb{R}}^{n_x\times m_1}$ and ${\mathbf{B}}_t\mathrm{\in }{\mathbb{R}}^{n_t\times m_2}$ are the evaluated univariate B-splines at the grid points $n_x$ and $n_t$ of the corresponding axes. The symbol $\otimes $ represents the Kronecker product of the matrix with the vector of ones having proper dimension and $\odot $ denotes the Hadamard product. Each column of $\mathbf{B}\mathrm{\in }{\mathbb{R}}^{n\mathrm{\times }m}$ can be reshaped into the unit ranked matrix and graphically displayed as a two-dimensional surface as shown in figure \ref{fig:S2}. The compact support of even multidimensional B-splines is evident from the figures as the values are nonzero within a small adjacent rectangular interval. It is conjectured that B-splines form about a set of nearly orthogonal basis functions \citep{berry2002bayesian} and the presence of many zeros in each of the evaluated functions are exploited to reduce the computational complexity and bring in numerical stability.

\subsection*{Closed-form expressions of optimum ADMM updates}

This section describes the derivation of the optimal solutions at each iterative update of \textbf{SNAPE}. The mathematical expressions of the iterative updates of the parameters depend on the form of the differential equation. Here, as an example, the iterative updates for the Burgers' equation are derived in detail.  For different ODEs or PDEs, the corresponding iterative updates can be computed following a similar approach. The Burgers' equation with field variable $u\left(x,t\right)$ has the following differential form,

\begin{equation} \label{SEQ04} 
	\frac{\partial u}{\partial t}+{\theta }_1u\frac{\partial u}{\partial x}+{\theta }_2\frac{{\partial }^2u}{\partial x^2}\ =\ 0 
\end{equation} 

The vector of noise corrupted measurement data $\mathbf{y}=u\left(x,t\right)+\boldsymbol{\epsilon }\boldsymbol{,\ \ }\mathbf{y}\boldsymbol{\ }\boldsymbol{\mathrm{\in }}{\mathbb{R}}^{n\times 1}$ where $n$\textbf{ }is the number of observations and $\boldsymbol{\epsilon }$\textbf{ }is i.i.d Gaussian noise with zero mean and unknown variance. \textbf{SNAPE} represents the PDE model and the associated parameter vector $\boldsymbol{\theta }=\left({\theta }_1,\ {\theta }_2\right)$ by expressing the process response $u\left(x,t\right)$ as an approximation to the linear combination of nonparametric basis functions given by

\begin{equation} \label{SEQ05} 
	u\left(x,t\right)\approx \overline{u}\left(x,t\right)=\sum^K_{k=1}{b_k\left(x,t\right){\beta }_k}={\mathbf{b}}^T\left(x,t\right)\boldsymbol{\beta } 
\end{equation} 

where $\mathbf{b}\left(x,t\right)=\{b_1\left(x,t\right),\dots ,b_K\left(x,t\right){\}}^T$ is the vector of basis functions and $\boldsymbol{\beta }={\left({\beta }_1,\dots ,{\beta }_K\right)}^T$ is the vector of basis coefficients. The basis functions ${\mathbf{b}}^T\left(x,t\right)$ are evaluated at $n$ observation points to obtain basis matrix $\overline{\mathbf{B}}\mathrm{\in }{\mathbb{R}}^{n\mathrm{\times }m}$, where $m$ is the number of columns in the basis matrix. The matrices corresponding to the linear terms of the PDE are evaluated as well.

\begin{equation} \label{SEQ06} 
	\begin{aligned}
		\frac{\mathrm{\partial }u}{\mathrm{\partial }t}\mathrm{\approx }{\left(\frac{\mathrm{\partial }\mathbf{b}\left(x,t\right)}{\mathrm{\partial }t}\right)}^T\boldsymbol{\beta }=&{\overline{\mathbf{B}}}_{\mathbf{0}}\boldsymbol{\beta } \\
		\frac{\partial u}{\partial x}\mathrm{\approx }{\left(\frac{\partial \mathbf{b}\left(x,t\right)}{\partial x}\right)}^T\boldsymbol{\beta }=&{\overline{\mathbf{B}}}_{\mathbf{1}}\boldsymbol{\beta } \\
		\frac{{\partial }^2u}{\partial x^2}\mathrm{\approx }{\left(\frac{{\partial }^2\mathbf{b}\left(x,t\right)}{\partial x^2}\right)}^T\boldsymbol{\beta }=&{\overline{\mathbf{B}}}_{\mathbf{2}}\boldsymbol{\beta }
	\end{aligned}
\end{equation}

where ${\overline{\mathbf{B}}}_{\mathbf{0}},{\overline{\mathbf{B}}}_{\mathbf{1}},{\overline{\mathbf{B}}}_{\mathbf{2}}\mathrm{\in }{\mathbb{R}}^{n\mathrm{\times }m}$ . The equivalent ADMM representation \citep{yang2011alternating} of the \textbf{SNAPE}'s optimization problem is given as

\begin{equation} \label{SEQ07} 
	\begin{array}{c}
		{\mathop{\mathrm{min}}_{\boldsymbol{\beta },{\theta }_1,{\theta }_2}\ \ \ \  \frac{1}{2}||\mathbf{y}-\overline{\mathbf{B}}\boldsymbol{\beta }{||}^2_2\ }+\frac{1}{2\mu }||\mathbf{r}{||}^2_2 \\[5pt] 
		subject\ to\ \ \ \ {\overline{\mathbf{B}}}_{\mathbf{0}}\boldsymbol{\beta }\boldsymbol{+}{\theta }_1\overline{\mathbf{B}}\boldsymbol{\beta }\boldsymbol{\bigodot }{\overline{\mathbf{B}}}_{\mathbf{1}}\boldsymbol{\beta }\boldsymbol{+}{\theta }_2{\overline{\mathbf{B}}}_{\mathbf{2}}\boldsymbol{\beta }\boldsymbol{+}\mathbf{r}\boldsymbol{=}\mathbf{0} \end{array}	
\end{equation}

where $\mathbf{r}\mathrm{\in }{\mathbb{R}}^n$ is an auxiliary variable. The scaled form of augmented Lagrangian of the above optimization problem is given as

\begin{equation} \label{SEQ08}
	\mathcal{L}\left(\beta ,{\theta }_1,{\theta }_2,u,r\right)=\frac{1}{2}||\mathbf{y}-\overline{\mathbf{B}}\boldsymbol{\beta }{||}^2_2+\frac{1}{2\mu }||\mathbf{r}{||}^2_2+\frac{\rho }{2}||G\left(\beta ,{\theta }_1,{\theta }_2,r\right)+\mathbf{u}{||}^2_2-\frac{\rho }{2}||\mathbf{u}{||}^2_2
\end{equation}

 where the function $G\left(\boldsymbol{\beta },{\theta }_1,{\theta }_2,\mathbf{r}\right)={\overline{\mathbf{B}}}_{\mathbf{0}}\boldsymbol{\beta }\boldsymbol{+}{\theta }_1\overline{\mathbf{B}}\boldsymbol{\beta }\boldsymbol{\bigodot }{\overline{\mathbf{B}}}_{\mathbf{1}}\boldsymbol{\beta }\boldsymbol{+}{\theta }_2{\overline{\mathbf{B}}}_{\mathbf{2}}\boldsymbol{\beta }\boldsymbol{+}\mathbf{r}$. The matrix$\boldsymbol{\mathrm{\ }}{\mathbf{B}}_{\mathbf{1}}\boldsymbol{\mathrm{=}}\overline{\mathbf{B}}\boldsymbol{\beta }\boldsymbol{\bigodot }{\overline{\mathbf{B}}}_{\mathbf{1}}$ is assumed constant for each iteration so that the function $G\left(\right)$ becomes linear in terms of $\boldsymbol{\beta }$. It is a biconvex relaxation of the original nonconvex problem when nonlinear differential equations are considered. The updates of the parameters at $k$${}^{th}$ step are computed by the following ADMM form \citep{yang2011alternating, boyd2011distributed}.
 
 \begin{equation} \label{SEQ09} 
 	\begin{aligned}
 		{\mathbf{r}}^{k+1}\coloneqq& \frac{\mathrm{\mu }\mathrm{\rho }}{1+\mathrm{\mu }\mathrm{\rho }}\left({\mathbf{u}}^k-G\left({\mathrm{\beta }}^k,{\theta }^k_1,{\theta }^k_2,{\mathbf{r}}^k\right)\right) \\
 		{\boldsymbol{\beta }}^{k+1}\coloneqq& \ \ \mathop{\mathrm{argmin}}_{\boldsymbol{\beta }}\left(\mathcal{L}\left({\boldsymbol{\beta }}^k,{\theta }^k_1,{\theta }^k_2,\mathbf{u},{\mathbf{r}}^{k+1}\right)\right) \\
 		{{\theta }_1}^{k+1}\coloneqq& \ \ \mathop{\mathrm{argmin}}_{{\theta }_1}\left(\mathcal{L}\left({\boldsymbol{\beta }}^{k+1},{\theta }^k_1,{\theta }^k_2,\mathbf{u},{\mathbf{r}}^{k+1}\right)\right) \\
 		{{\theta }_2}^{k+1}\coloneqq& \ \ \mathop{\mathrm{argmin}}_{{\theta }_2}\left(\mathcal{L}\left({\boldsymbol{\beta }}^{k+1},{{\theta }_1}^{k+1},{\theta }^k_2,\mathbf{u},{\mathbf{r}}^{k+1}\right)\right) \\
 		{\mathbf{u}}^{k+1}\coloneqq& {\mathbf{u}}^k+\gamma \left(G\left({\boldsymbol{\beta }}^{k+1},{{\theta }_1}^{k+1},{{\theta }_2}^{k+1},{\mathbf{r}}^{k+1}\right)\right)
 	\end{aligned}
 \end{equation}

Each iterative update of the parameters involves optimization of the Lagrangian for the corresponding parameter. The optimum values ${\boldsymbol{\beta }}^*$, ${\theta }^*_1$, and ${\theta }^*_2$ for each ADMM iteration step is computed by optimizing the following loss function

\begin{equation} \label{SEQ10} 
	\begin{aligned}
		J=&\frac{1}{2}||\mathbf{y}-\overline{\mathbf{B}}\boldsymbol{\beta }{||}^2_2+\frac{1}{2\mu }||\mathbf{r}{||}^2_2+\frac{\mathrm{\rho }}{2}||{\overline{\mathbf{B}}}_{\mathbf{0}}\boldsymbol{\beta }\boldsymbol{+}{\theta }_1{\mathbf{B}}_{\mathbf{1}}\boldsymbol{\beta }\boldsymbol{+}{\theta }_2{\overline{\mathbf{B}}}_{\mathbf{2}}\boldsymbol{\beta }+(\mathbf{r}+\mathbf{u}){||}^2_2-\frac{\rho }{2}||\mathbf{u}{||}^2_2 \\
		J=&\frac{1}{2}{\left(\mathbf{y}-\overline{\mathbf{B}}\boldsymbol{\beta }\right)}^T\left(\mathbf{y}-\overline{\mathbf{B}}\boldsymbol{\beta }\right)+\frac{1}{2\mu }{\mathbf{r}}^T\mathbf{r}\boldsymbol{+}\frac{\rho }{2}{\left({\overline{\mathbf{B}}}_{\mathbf{0}}\boldsymbol{\beta }\boldsymbol{+}{\theta }_1{\mathbf{B}}_{\mathbf{1}}\boldsymbol{\beta }\boldsymbol{+}{\theta }_2{\overline{\mathbf{B}}}_{\mathbf{2}}\boldsymbol{\beta }+\left(\mathbf{r}+\mathbf{u}\right)\right)}^T\left({\overline{\mathbf{B}}}_{\mathbf{0}}\boldsymbol{\beta }\boldsymbol{+}{\theta }_1{\mathbf{B}}_{\mathbf{1}}\boldsymbol{\beta }\boldsymbol{+}{\theta }_2{\overline{\mathbf{B}}}_{\mathbf{2}}\boldsymbol{\beta }+\left(\mathbf{r}+\mathbf{u}\right)\right) \\
		&-\frac{\rho }{2}{\mathbf{u}}^T\mathbf{u} \\
		J=&\frac{1}{2}\left({\mathbf{y}}^{\mathbf{T}}\mathbf{y}-2{\boldsymbol{\beta }}^{\mathbf{T}}{\overline{\mathbf{B}}}^{\mathbf{T}}\mathbf{y}\boldsymbol{+}{\boldsymbol{\beta }}^{\mathbf{T}}{\overline{\mathbf{B}}}^{\mathbf{T}}\overline{\mathbf{B}}\boldsymbol{\beta }\right)+\frac{1}{2\mu }{\mathbf{r}}^T\mathbf{r} \\
		&\boldsymbol{+}\frac{\rho }{2}\left({{\boldsymbol{\beta }}^{\mathbf{T}}{\overline{\mathbf{B}}}^{\mathbf{T}}_{\mathbf{0}}\overline{\mathbf{B}}}_{\mathbf{0}}\boldsymbol{\beta }\boldsymbol{+}{2\theta }_1{\boldsymbol{\beta }}^{\mathbf{T}}{\overline{\mathbf{B}}}^{\mathbf{T}}_{\mathbf{0}}{\mathbf{B}}_{\mathbf{1}}\boldsymbol{\beta }\boldsymbol{+}{2\theta }_2{\boldsymbol{\beta }}^{\mathbf{T}}{\overline{\mathbf{B}}}^{\mathbf{T}}_{\mathbf{0}}{\overline{\mathbf{B}}}_{\mathbf{2}}\boldsymbol{\beta }\boldsymbol{+}{\theta }^2_1{\boldsymbol{\beta }}^{\mathbf{T}}{\mathbf{B}}^T_{\mathbf{1}}{\mathbf{B}}_{\mathbf{1}}\boldsymbol{\beta }\boldsymbol{+}{\theta }^2_2{\boldsymbol{\beta }}^{\mathbf{T}}{\overline{\mathbf{B}}}^T_{\mathbf{2}}{\overline{\mathbf{B}}}_{\mathbf{2}}\boldsymbol{\beta }\boldsymbol{+}{2\theta }_1{\theta }_2{\boldsymbol{\beta }}^{\mathbf{T}}{\mathbf{B}}^T_{\mathbf{1}}{\overline{\mathbf{B}}}_{\mathbf{2}}\boldsymbol{\beta }\right) \\
		&\boldsymbol{+}\frac{\rho }{2}\left(2({\boldsymbol{\beta }}^{\mathbf{T}}{\overline{\mathbf{B}}}^{\mathbf{T}}_{\mathbf{0}}\boldsymbol{+}{\theta }_1{\boldsymbol{\beta }}^{\mathbf{T}}{\mathbf{B}}^T_{\mathbf{1}}+{\theta }_2{\boldsymbol{\beta }}^{\mathbf{T}}{\overline{\mathbf{B}}}^T_{\mathbf{2}})\left(\mathbf{r}+\mathbf{u}\right)\right)-\frac{\rho }{2}{\mathbf{u}}^T\mathbf{u}
	\end{aligned}
\end{equation}

The gradient of this loss function with respect to $\boldsymbol{\beta }$ is given as:

\begin{equation} \label{SEQ11} 
	\begin{aligned}
		\frac{\mathrm{\partial }J}{\mathrm{\partial }\boldsymbol{\beta }}=&\frac{1}{2}\left(\boldsymbol{2}{\overline{\mathbf{B}}}^{\mathbf{T}}\overline{\mathbf{B}}\boldsymbol{\beta }-2{\overline{\mathbf{B}}}^{\mathbf{T}}\mathbf{y}\right) \\
		&\boldsymbol{+}\frac{\rho }{2}\left({\mathbf{2}{\overline{\mathbf{B}}}^{\mathbf{T}}_{\mathbf{0}}\overline{\mathbf{B}}}_{\mathbf{0}}\mathbf{\beta }\boldsymbol{+}{4\theta }_1{\overline{\mathbf{B}}}^{\mathbf{T}}_{\mathbf{0}}{\mathbf{B}}_{\mathbf{1}}\boldsymbol{\beta }\boldsymbol{+}{4\theta }_2{\overline{\mathbf{B}}}^{\mathbf{T}}_{\mathbf{0}}{\overline{\mathbf{B}}}_{\mathbf{2}}\boldsymbol{\beta }\boldsymbol{+}{2\theta }^2_1{\mathbf{B}}^T_{\mathbf{1}}{\mathbf{B}}_{\mathbf{1}}\boldsymbol{\beta }\boldsymbol{+}{2\theta }^2_2{\overline{\mathbf{B}}}^T_{\mathbf{2}}{\overline{\mathbf{B}}}_{\mathbf{2}}\boldsymbol{\beta }\boldsymbol{+}{4\theta }_1{\theta }_2{\mathbf{B}}^T_{\mathbf{1}}{\overline{\mathbf{B}}}_{\mathbf{2}}\boldsymbol{\beta }\right) \\
		&\boldsymbol{+}\frac{\rho }{2}\left(2\left({\overline{\mathbf{B}}}^{\mathbf{T}}_{\mathbf{0}}\boldsymbol{+}{\theta }_1{\mathbf{B}}^T_{\mathbf{1}}+{\theta }_2{\overline{\mathbf{B}}}^T_{\mathbf{2}}\right)\left(\mathbf{r}+\mathbf{u}\right)\right)
	\end{aligned}
\end{equation}

The closed-form expression for the optimum parameter ${\boldsymbol{\beta }}^*$ is obtained by equating $\frac{\mathrm{\partial }J}{\mathrm{\partial }\boldsymbol{\beta }}=0$.

\begin{equation} \label{SEQ12}
	\begin{aligned} 
	{\boldsymbol{\beta }}^*=&{[{\overline{\mathbf{B}}}^{\mathbf{T}}\overline{\mathbf{B}}+\rho ({{\overline{\mathbf{B}}}^{\mathbf{T}}_{\mathbf{0}}\overline{\mathbf{B}}}_{\mathbf{0}}\boldsymbol{+}{2\theta }_1{\overline{\mathbf{B}}}^{\mathbf{T}}_{\mathbf{0}}{\mathbf{B}}_{\mathbf{1}}\boldsymbol{+}{2\theta }_2{\overline{\mathbf{B}}}^{\mathbf{T}}_{\mathbf{0}}{\overline{\mathbf{B}}}_{\mathbf{2}}\boldsymbol{+}{\theta }^2_1{\mathbf{B}}^T_{\mathbf{1}}{\mathbf{B}}_{\mathbf{1}}\boldsymbol{+}{\theta }^2_2{\overline{\mathbf{B}}}^T_{\mathbf{2}}{\overline{\mathbf{B}}}_{\mathbf{2}}\boldsymbol{+}{2\theta }_1{\theta }_2{\mathbf{B}}^T_{\mathbf{1}}{\overline{\mathbf{B}}}_{\mathbf{2}})]}^{-1} \\
	&[{\overline{\mathbf{B}}}^{\mathbf{T}}\mathbf{y}-\rho \left({\overline{\boldsymbol{B}}}^{\boldsymbol{T}}_{\boldsymbol{0}}\boldsymbol{+}{\theta }_1{\mathbf{B}}^T_{\mathbf{1}}+{\theta }_2{\overline{\mathbf{B}}}^T_{\mathbf{2}}\right)\left(\mathbf{r}+\mathbf{u}\right)] 
\end{aligned}
\end{equation}

Similarly, the gradient of the loss function with respect to ${\theta }_1$ is given as:

\begin{equation} \label{SEQ13}
	\frac{\mathrm{\partial }J}{\mathrm{\partial }{\theta }_1}=\frac{\rho }{2}\left(2{\boldsymbol{\beta }}^{\mathbf{T}}{\overline{\mathbf{B}}}^{\mathbf{T}}_{\mathbf{0}}{\mathbf{B}}_{\mathbf{1}}\boldsymbol{\beta }\boldsymbol{+}2{\theta }_1{\boldsymbol{\beta }}^{\mathbf{T}}{\mathbf{B}}^T_{\mathbf{1}}{\mathbf{B}}_{\mathbf{1}}\boldsymbol{\beta }\boldsymbol{+}2{\theta }_2{\boldsymbol{\beta }}^{\mathbf{T}}{\mathbf{B}}^T_{\mathbf{1}}{\overline{\mathbf{B}}}_{\mathbf{2}}\boldsymbol{\beta }\boldsymbol{+}2{\boldsymbol{\beta }}^{\mathbf{T}}{\mathbf{B}}^T_{\mathbf{1}}\left(\mathbf{r}+\mathbf{u}\right)\right)
\end{equation}

The closed-form expression for the optimum parameter ${\theta }^*_1$ is obtained by equating $\frac{\mathrm{\partial }J}{\mathrm{\partial }{\theta }_1}=0$.

\begin{figure}
	\centering
	\includegraphics[width=0.8\textwidth]{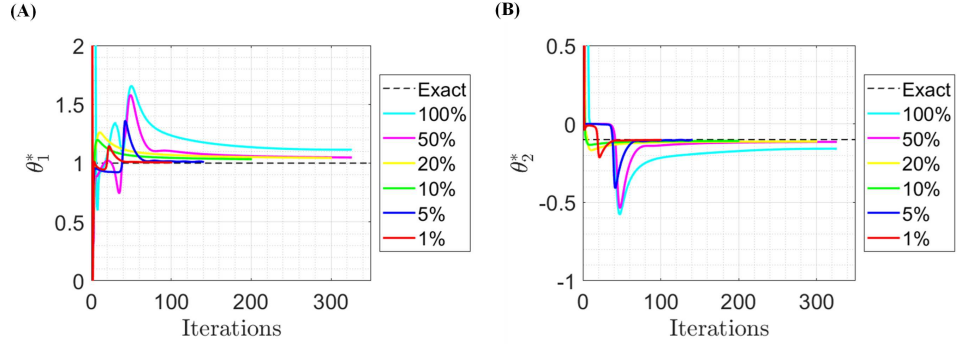}
	\caption{Convergence plot of the parameters of the Burgers' equation. The black dashed line represents the exact value of the parameter used to simulate the response. The colored lines represent the updated parameter value at each of the iteration steps corresponding to the percentage of Gaussian noise corrupted measured data.}\label{fig:S3}
\end{figure}

\begin{equation} \label{SEQ14} 
	{\theta }^*_1=-{[{\boldsymbol{\beta }}^{\mathbf{T}}{\mathbf{B}}^T_{\mathbf{1}}{\mathbf{B}}_{\mathbf{1}}\boldsymbol{\beta }]}^{-1}[{\boldsymbol{\beta }}^{\mathbf{T}}{\overline{\mathbf{B}}}^{\mathbf{T}}_{\mathbf{0}}{\mathbf{B}}_{\mathbf{1}}\boldsymbol{\beta }\boldsymbol{+}{\theta }_2{\boldsymbol{\beta }}^{\mathbf{T}}{\mathbf{B}}^T_{\mathbf{1}}{\overline{\mathbf{B}}}_{\mathbf{2}}\boldsymbol{\beta }\boldsymbol{+}{\boldsymbol{\beta }}^{\mathbf{T}}{\mathbf{B}}^T_{\mathbf{1}}\left(\mathbf{r}+\mathbf{u}\right)] 
\end{equation}

Similarly, the closed-form expression for the optimum parameter ${\theta }^*_2$ is obtained by equating $\frac{\mathrm{\partial }J}{\mathrm{\partial }{\theta }_2}=0$.

\begin{equation} \label{SEQ15} 
	{\theta }^*_2=-{[{\boldsymbol{\beta }}^{\mathbf{T}}{\mathbf{B}}^T_{\mathbf{2}}{\mathbf{B}}_{\mathbf{2}}\boldsymbol{\beta }]}^{-1}[{\boldsymbol{\beta }}^{\mathbf{T}}{\overline{\mathbf{B}}}^{\mathbf{T}}_{\mathbf{0}}{\mathbf{B}}_{\mathbf{2}}\boldsymbol{\beta }\boldsymbol{+}{\theta }_1{\boldsymbol{\beta }}^{\mathbf{T}}{\mathbf{B}}^T_{\mathbf{2}}{\overline{\mathbf{B}}}_{\mathbf{1}}\boldsymbol{\beta }\boldsymbol{+}{\boldsymbol{\beta }}^{\mathbf{T}}{\mathbf{B}}^T_{\mathbf{2}}\left(\mathbf{r}+\mathbf{u}\right)] 
\end{equation}

The following algorithm demonstrates the parameter estimation of the Burgers' equation model using \textbf{SNAPE}. \vspace{5mm}

\begin{center}
\begin{tabular}{p{3.5in}} \hline 
	\\[2pt]
	\hspace{15mm} Algorithm: \textbf{SNAPE} \textbf{(\textit{Burgers' Equation})} \\[7pt] \hline 
	\\[3pt]
	\hspace{5mm} Initialize $\rho \mathrm{>0}$, $\mu \mathrm{>0}$, $\gamma \mathrm{>0}$, ${\theta }^0_1$, \textbf{and} ${\theta }^0_2$ \\
	 \hspace{5mm} $\mathbf{u}^0\mathrm{\leftarrow }0$ \\
	 \hspace{5mm} $\mathbf{r}^0\mathrm{\leftarrow }0$ \\
	 \hspace{5mm} ${\boldsymbol{\beta } }^0\mathrm{\leftarrow }{[{\overline{\mathbf{B}}}^T\overline{\mathbf{B}}]}^{-1}[{\overline{\mathbf{B}}}^T\mathbf{y}]$ \\
	 \hspace{5mm} $k\mathrm{\leftarrow }0$ \\
	 \hspace{5mm} while \textbf{\textit{till convergence}} do \\
	 \hspace{10mm} $\mathrm{\mathbf{B}}\mathrm{\leftarrow }\overline{\mathbf{B}}\beta \bigodot {\overline{\mathbf{B}}}_1$ \\ 
	 \hspace{10mm} $\mathbf{r}^{k+1}\mathrm{\leftarrow }\frac{\mu \rho }{1+\mu \rho }\left(\mathbf{u}^k-G\left({\boldsymbol{\beta } }^k,{\theta }^k_1,{\theta }^k_2,\mathbf{r}^k\right)\right)$ \\ 
	 \hspace{10mm} ${\boldsymbol{\beta } }^{k+1}\mathrm{\leftarrow }{\boldsymbol{\beta } }^*$ \\
	 \hspace{10mm} ${{\theta }_1}^{k+1}\mathrm{\leftarrow }{\theta }^*_1$ \\
	 \hspace{10mm} ${{\theta }_2}^{k+1}\mathrm{\leftarrow }{\theta }^*_2$ \\
	 \hspace{10mm} $\mathbf{u}^{k+1}\mathrm{\leftarrow }\mathbf{u}^k+\gamma \left(G\left({\boldsymbol{\beta } }^{k+1},{{\theta }_1}^{k+1},{{\theta }_2}^{k+1},\mathbf{r}^{k+1}\right)\right)$ \\
	 \hspace{10mm} $k\mathrm{\leftarrow }k+1$  \\[5pt] \hline 
\end{tabular} \\[20pt]
\end{center}

\begin{figure}
	\centering
	\includegraphics[width=0.8\textwidth]{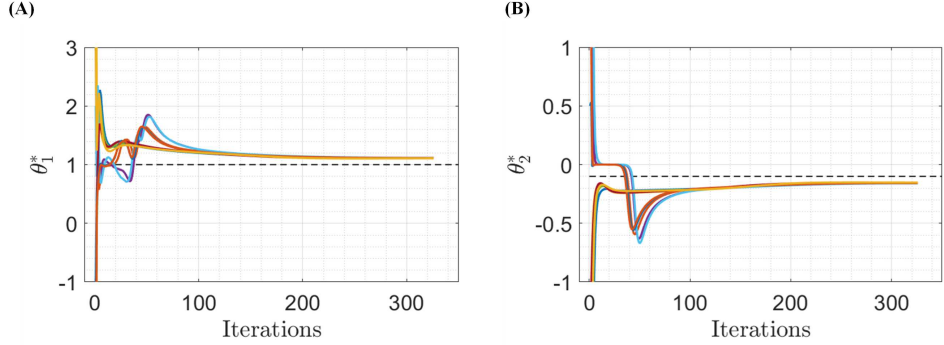}
	\caption{ Convergence plot of the parameters of the Burgers' equation where each colored path is corresponding to a random initialization from the uniform distributions ${\theta }^0_1\sim U\left(-10,10\right)$ and ${\theta }^0_2\sim U\left(-10,10\right)$. The black dashed line represents the exact value of the parameter used to simulate the response. For all the initializations, the measured response is corrupted with an extreme level (100\%) of Gaussian noise.}\label{fig:S4}
\end{figure}

The figure \ref{fig:S3} shows the plots of each of the optimum parameters of Burgers' equation for each iteration step of \textbf{SNAPE}. The figure demonstrates the convergence of \textbf{SNAPE} for measured data corrupted with low (1\%) to extreme (100\%) levels of Gaussian noise. As expected, with increasing noise content, \textbf{SNAPE} requires more iterations to reach convergence. In this example, the initial value of the parameter is set to ${\theta }^0_1=3.0$ and ${\theta }^0_2=3.0$.

The convergence to the optimum parameter values of the model does not depend on the initialization of the model's parameters. \textbf{SNAPE} exhibits insensitivity towards the choice of ${\theta }^0_1$ and ${\theta }^0_2$. Figure \ref{fig:S4} shows the convergence plots of the parameters of Burgers' equation using \textbf{SNAPE} for 10 different initializations randomly sampled from the uniform distributions ${\theta }^0_1\sim U\left(-10,10\right)$ and ${\theta }^0_2\sim U\left(-10,10\right)$. The original data is corrupted with 100\% noise for all the random instances of initialization to inspect the algorithm's convergence stability under extreme perturbation. 

\subsection*{Examples}

This section describes the theory-guided learning of the Korteweg-de Vries equation and the Kuramoto-Sivashinsky equation from its noise-corrupted measured using \textbf{SNAPE}. The performance of the parameter estimation is already demonstrated in Table 1. The simulated data for both models is obtained from \citet{rudy2017data}.

\subsubsection*{Korteweg-de Vries (KdV) equation}

\begin{figure}
	\centering
	\includegraphics[width=1.0\textwidth]{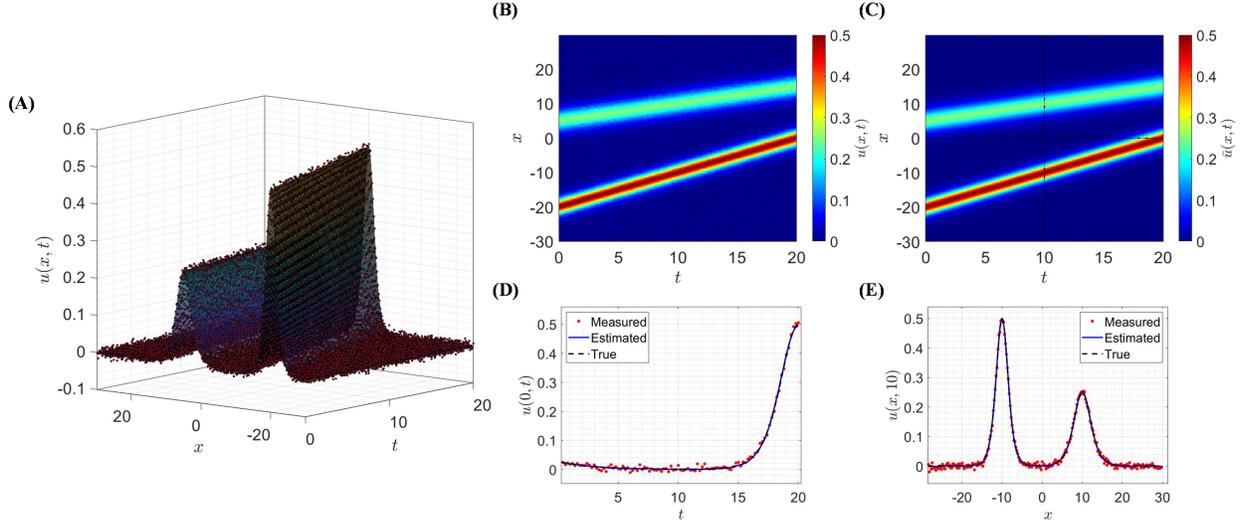}
	\caption{\textbf{(A)} 5\% Gaussian noise corrupted measured data points overlaid on the surface of the true solution of the Korteweg-de Vries equation.\textbf{(B)} The measured response shows two traveling waves with different amplitudes.\textbf{(C)} The analytical approximate solution of the underlying PDE model. The black dashed lines indicate the position and time instant of response shown as 1D plots in the figures below. The comparison of the estimated solution from the noisy measured data to the true solution \textbf{(D)} at position$\ x=0$ and \textbf{(E)} at time instant $t=10$ reveals the efficacy and robustness of SNAPE.}\label{fig:S5}
\end{figure}

The KdV equation has relations to many physical problems including but not limited to waves in shallow water with weakly nonlinear restoring force and acoustic waves in plasma or on a crystal lattice. The corresponding PDE model is given as

\begin{equation} \label{SEQ16} 
	\frac{\mathrm{\partial }u}{\mathrm{\partial }t}+{\theta }_1u\frac{\mathrm{\partial }u}{\mathrm{\partial }x}+{\theta }_2\frac{{\mathrm{\partial }}^3u}{\mathrm{\partial }x^3}=0 
\end{equation} 

The numerical simulation of the response $u\left(x,t\right)\mathrm{\in }{\mathbb{R}}^{512\times 201}$ is performed in the domain $x\mathrm{\in }\left[-30,\ 30\right]$ and $t\mathrm{\in }\left[0,\ 20\right]$ for the parameter values $\boldsymbol{\theta }=\left(6.0,\ 1.0\right)$. It models 1D wave propagation of two non-interacting traveling waves of different amplitudes. As shown in Table \ref{tab:01}, \textbf{SNAPE} robustly estimates the parameters of the KdV equation with high accuracy for cases where the simulated response is corrupted with 1\% and 5\% Gaussian noise. Figure \ref{fig:S5} (A) shows one such instance of measured data corrupted with 5\% noise overlaid on the true response of the KdV equation. The estimated functional solution approximates well the true response of the model as shown in the time history plot in figure \ref{fig:S5} (D) and an instantaneous snapshot of response in figure \ref{fig:S5} (D).

\subsubsection*{Kuramoto-Sivashinsky (KS) equation}

\begin{figure}
	\centering
	\includegraphics[width=1.0\textwidth]{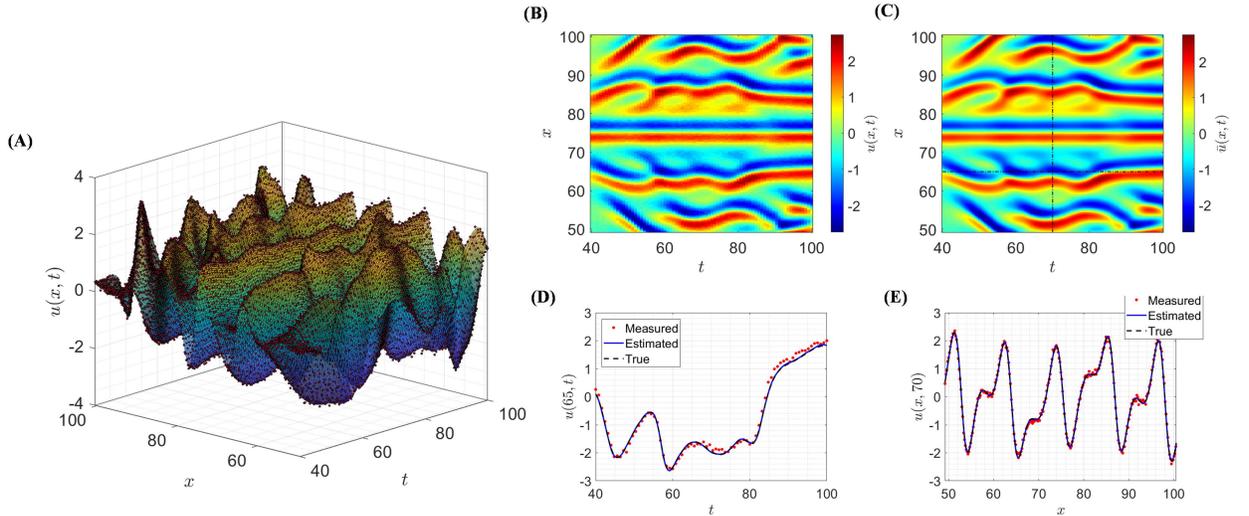}
	\caption{\textbf{(A)} The measured data points with 5\% Gaussian noise overlaid on the surface of the true solution of the Kuramoto-Sivashinsky equation.\textbf{(B)} The measured response demonstrates a complex spatiotemporal pattern.\textbf{(C)} The analytical approximate solution of the underlying PDE model. The black dashed lines indicate the position and time instant of response shown as 1D plots in the figures below. The comparison of the estimated solution from the noisy measured data to the true solution \textbf{(D)} at position$\ x=65$ and \textbf{(E)} at time instant $t=70$ reveals the efficacy and robustness of SNAPE.}\label{fig:S6}
\end{figure}

The fourth-order nonlinear PDE of the KS equation has attracted a great deal of attention to model complex spatiotemporal dynamics of spatially extended systems that are driven far from equilibrium by intrinsic instabilities such as instabilities in laminar flame fonts, phase dynamics in reaction-diffusion systems, and instabilities of dissipative trapped ion modes in plasmas. The PDE model of the KS equation in one space dimension is given as

\begin{equation} \label{SEQ17} 
	\frac{\mathrm{\partial }u}{\mathrm{\partial }t}+{\theta }_1u\frac{\mathrm{\partial }u}{\mathrm{\partial }x}+{\theta }_2\frac{{\mathrm{\partial }}^2u}{\mathrm{\partial }x^2}+{\theta }_3\frac{{\mathrm{\partial }}^4u}{\mathrm{\partial }x^4}=0 
\end{equation} 

The original data consists of solution domain $x\mathrm{\in }\left[0,\ 100.5\right]$ and $t\mathrm{\in }\left[0,\ 100\right]$ for the parameter values $\boldsymbol{\theta }=\left(1.0,\ 1.0,\ 1.0\right)$. But in the present study, a part of the response $u\left(x,t\right)\mathrm{\in }{\mathbb{R}}^{524\times 151}$ in the domain $x\mathrm{\in }\left[49.2,\ 100.5\right]$ and $t\mathrm{\in }\left[40,\ 100\right]$ is used to infer the parameters of the model. Even though the model consists of a fourth-order derivative and the measured response is corrupted with Gaussian noise (1\% and 5\%), \textbf{SNAPE} is successful in estimating the parameters with reasonable accuracy and uncertainty as tabulated in Table \ref{tab:01}. Figure \ref{fig:S6} (A) shows one such instance of measured data corrupted with 5\% noise overlaid on the true response of the KS equation. The estimated analytical solution approximates well the true response of the model as shown in the time history plot in figure \ref{fig:S6} (D) and an instantaneous snapshot of response in figure \ref{fig:S6} (D).

\subsubsection*{Comparative study}

\begin{table}[htbp]
	\centering
	\caption{\textbf{Comparative performance of SNAPE. }The relative error of parameter estimation along with its variance in percentage for \textbf{SNAPE }is compared with that of \citet{rudy2017data} for the following PDE models with the same dataset. In general, the accuracy and robustness of \textbf{SNAPE}'s estimation are better from 5\% Gaussian noise corrupted data than that of \citet{rudy2017data} from data with 1\% Gaussian added noise.}
	\begin{tabular}{|p{1.0in}|p{1.8in}|p{0.8in}|p{0.7in}|p{0.7in}|} \hline 
		{} & {} & {} & {} & {} \\[2pt]
		\textbf{Differential Equations} & \hspace{15mm} \textbf{Form} & \textbf{\citet{rudy2017data}\newline (1\% Noise)} & \textbf{SNAPE \newline (1\% Noise)} & \textbf{SNAPE \newline (5\% Noise)} \\ \hline 
		{} & {} & {} & {} & {} \\[2pt]
		\textbf{Kuramoto-Sivashinsky equation} & $u_t+{\theta }_1uu_x+{\theta }_2u_{xx}+{\theta }_3u_{xxxx}=0$  & $52\pm 1.4\%$ & $3.6\pm 0.92\%$\textbf{\textit{\newline }}  & $20.8\pm 19\%$\newline  \\ \hline 
		{} & {} & {} & {} & {} \\[2pt]
		\textbf{Burgers' equation} & $u_t+{\theta }_1uu_x+{\theta }_2u_{xx}=0$\textit{ } & $0.8\pm 0.6\%$ & $1.0\pm 0.08\%$ & $1.0\pm 0.55\%$\newline  \\ \hline 
		{} & {} & {} & {} & {} \\[2pt]
		\textbf{Korteweg-de Vries equation} & $u_t+{\theta }_1uu_x+{\theta }_2u_{xxx}=0$  & $7\pm 5\%$ & $0.4\pm 0.06\%$ & $0.7\pm 0.28\%$\newline  \\ \hline 
		{} & {} & {} & {} & {} \\[2pt]
		\textbf{Nonlinear Schr\"{o}dinger equation} & ${\psi }_t+{\theta }_1{\psi }_{xx}+{\theta }_2{\left|\psi \right|}^2\psi =0$\textit{ } & $3\pm 1\%$ & $0.9\pm 0.13\%$ & $5.7\pm 0.31\%$ \\ \hline 
	\end{tabular}
	\label{tab:S01}%
\end{table}%

\begin{table}[htbp]
	\centering
	\caption{\textbf{Comparative performance of SNAPE for Navier-Stokes equation. }The relative error of parameter estimation along with its variance in percentage for \textbf{SNAPE }is compared with that of \citep{raissi2019physics} with the same dataset. The accuracy of \textbf{SNAPE}'s estimation from 5\% Gaussian noise corrupted data is comparable to that of \citep{raissi2019physics} from data with 1\% Gaussian added noise.}
	\begin{tabular}{|p{1.0in}|p{1.8in}|p{0.8in}|p{0.7in}|p{0.7in}|} \hline 
		{} & {} & {} & {} & {} \\[2pt]
		\textbf{Differential Equations} & \hspace{15mm} \textbf{Form} & \textbf{\citet{raissi2019physics} \newline  (1\% Noise)} & \textbf{SNAPE                                  (1\% Noise)} & \textbf{SNAPE                                  (5\% Noise)} \\ \hline 
		{} & {} & {} & {} & {} \\[2pt]
		\textbf{Navier-Stokes equation} & ${\omega }_t+{\theta }_1{\omega }_{xx}+{\theta }_2{\omega }_{yy}+{\theta }_3u{\omega }_x+{\theta }_4v{\omega }_y=0$\textit{ } & $8.9\%$ & $9.1\pm 0.07\%$\textbf{\textit{\newline }}  & $9.2\pm 1.6\%$\newline  \\ \hline 
	\end{tabular}
	\label{tab:S02}%
\end{table}%

This section compares the efficacy of the proposed method of \textbf{SNAPE} with that of the prevalent methods in the literature of estimating parameters of PDE models. The data for the PDE models of KS equation, Burgers' equation, KdV equation, and NLSE are obtained from the same source of \citet{rudy2017data} whose results are compared with \textbf{SNAPE} in Table \ref{tab:S01}. The same data for the estimation provides a common basis for the comparison. The regression-based method in \citet{rudy2017data} is demonstrated for measurement noise up to 1\%. However, the accuracy and robustness of \textbf{SNAPE} not only outperforms that of \citet{rudy2017data} for all the PDE models corrupted with 1\% Gaussian noise, but also performs better with 5\% added noise for almost all the cases.
	
The velocity field data for the Navier-Stokes equation is obtained from \citet{raissi2019physics}. The vorticity field data is numerically obtained from it and subsequently, the two velocity components and vorticity field data are corrupted with Gaussian noise to replicate the measurement noise. The following table compares the performance of \textbf{SNAPE} with the deep learning-based method in \citet{raissi2019physics} for the same dataset. In \citet{raissi2019physics} the authors estimate the parameters from one random instance of added noise, but here the robustness and repeatability of \textbf{SNAPE} are demonstrated by performing parameter estimation from 10 bootstrap samples of noise-induced data. The accuracy of estimation using \textbf{SNAPE }for 5\% noise shown in Table \ref{tab:S02} is comparable to that in \citet{raissi2019physics} for 1\% noise.

\bibliographystyle{unsrtnat}
\bibliography{references}  






\end{document}